\documentclass[conference]{IEEEtran}
\usepackage{times}

\usepackage[numbers]{natbib}
\usepackage{multicol}
\usepackage[bookmarks=true,colorlinks=true,urlcolor=Red,citecolor=Green]{hyperref}
\usepackage[dvipsnames]{xcolor}
\usepackage{amsmath}
\usepackage{amssymb}
\usepackage{mathtools}
\usepackage{siunitx}
\usepackage{graphicx}
\usepackage{adjustbox}
\usepackage{float}
\usepackage{wrapfig}
\usepackage[font=small]{subcaption}
\usepackage[font=small,skip=3pt,tableposition=top]{caption}
\usepackage{booktabs}
\usepackage{threeparttable}
\usepackage{array,multirow}
\usepackage{microtype}
\usepackage{soul}
\usepackage{balance}

\def\figurespace{\vspace{-2.8ex}}

\newcommand{\comment}[1]{}
\newcommand\bevnet{TerrainNet}

\makeatletter
\def\blfootnote{\gdef\@thefnmark{}\@footnotetext}
\makeatother

\renewcommand{\baselinestretch}{0.98}

\begin{document}

% \title{\bevnet{}: Visual Terrain Modeling for High-speed, Off-road Navigation}

\title{\bevnet{}: Visual Modeling of Complex Terrain for High-speed, Off-road Navigation}

% \title{\bevnet{}: Visual Terrain Modeling for High-speed Navigation on Complex Off-road Terrains}

% You will get a Paper-ID when submitting a pdf file to the conference system
% \author{
% Author Names Omitted for Anonymous Review. Paper-ID 146 \\ \href{https://sites.google.com/view/visual-terrain-modeling}{\texttt{https://sites.google.com/view/visual-terrain-modeling}}}

\author{\authorblockN{Xiangyun Meng\textsuperscript{1}, Nathan Hatch\textsuperscript{1}, Alexander Lambert\textsuperscript{1}, Anqi Li\textsuperscript{1}, Nolan Wagener\textsuperscript{2}, Matthew Schmittle\textsuperscript{1}, JoonHo Lee\textsuperscript{1},\\
Wentao Yuan\textsuperscript{1}, Zoey Chen\textsuperscript{1}, Samuel Deng\textsuperscript{1}, Greg Okopal\textsuperscript{1}, Dieter Fox\textsuperscript{1}, Byron Boots\textsuperscript{1}, Amirreza Shaban\textsuperscript{1}}
\textsuperscript{1}University of Washington \hspace{5mm} \textsuperscript{2}Georgia Institute of Technology\\
\href{https://sites.google.com/view/visual-terrain-modeling}{\texttt{https://sites.google.com/view/visual-terrain-modeling}}
}

% avoiding spaces at the end of the author lines is not a problem with
% conference papers because we don't use \thanks or \IEEEmembership

%Homer Simpson\authorrefmark{2},
%Montgomery Scott\authorrefmark{3} and
%Eldon Tyrell\authorrefmark{4}}
%Telephone: (800) 555--1212, Fax: (888) 555--1212}
%\authorblockA{\authorrefmark{4}Tyrell Inc., 123 Replicant Street, Los Angeles, California 90210--4321}}

\maketitle

\begin{abstract}
Effective use of camera-based vision systems is essential for robust performance in autonomous off-road driving, particularly in the high-speed regime. Despite success in structured, on-road settings, current end-to-end approaches for scene prediction have yet to be successfully adapted for complex outdoor terrain. To this end, we present \bevnet{}, a vision-based terrain perception system for \textit{semantic} and \textit{geometric} terrain prediction for aggressive, off-road navigation. The approach relies on several key insights and practical considerations for achieving reliable terrain modeling. The network includes a multi-headed output representation to capture fine- and coarse-grained terrain features necessary for estimating traversability. Accurate depth estimation is achieved using self-supervised depth completion with multi-view RGB and stereo inputs. Requirements for real-time performance and fast inference speeds are met using efficient, learned image feature projections. Furthermore, the model is trained on a large-scale, real-world off-road dataset collected across a variety of diverse outdoor environments. We show how \bevnet{} can also be used for costmap prediction and provide a detailed framework for integration into a planning module. We demonstrate the performance of \bevnet{} through extensive comparison to current state-of-the-art baselines for camera-only scene prediction. Finally, we showcase the effectiveness of integrating \bevnet{} within a complete autonomous-driving stack by conducting a real-world vehicle test in a challenging off-road scenario. 
\end{abstract}

\IEEEpeerreviewmaketitle

% commented out for anonymized submission
\blfootnote{\hspace{-1em}Distribution Statement A. Approved for Public Release, Distribution Unlimited.}

\section{Introduction}
\begin{figure}[t]
    \centering
    \includegraphics[width=1\columnwidth]{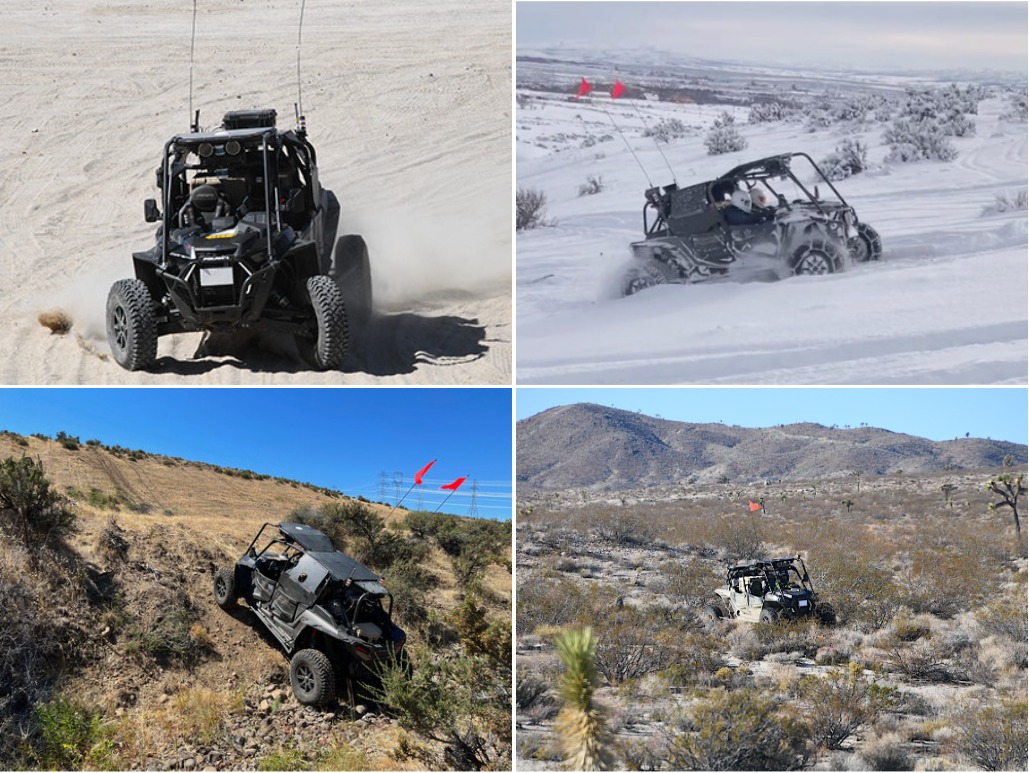}
    \caption{High-speed driving in complex off-road environments requires joint reasoning of terrain semantics and geometry. Top row: a vehicle can drive at high-speed on a dirt road but has to be more cautious in snow due to wheel slipping. Bottom row: a vehicle needs to estimate terrain slopes and sizes of vegetation for safe planning and control.}
    \vspace{-5mm}
    \label{fig:overview}
\end{figure}

Autonomous robot navigation in off-road environments has seen a wide range of applications including search and rescue~\citep{ventura2012search}, agriculture~\citep{duckett2018agricultural}, planetary exploration~\citep{thoesen2021planetary,wilcox1992robotic}, and defense~\citep{krotkov1999defense}. Unlike indoor or on-road environments where traversable areas and non-traversable areas are clearly separated, off-road terrains exhibit a wide range of traversability that require a comprehensive understanding of the semantics and geometry of the terrain (Figure~\ref{fig:overview}).

Current off-road navigation systems typically rely on LiDAR to obtain a 3D point cloud of the environment for semantic and geometric analysis~\cite{guan2022tns,maturana2018real,shaban2022semantic,silver2010learning,stolzle2022reconstructing}. While LiDAR sensors provide accurate spatial information, the resulting point cloud is rather sparse, making it tricky to build a \emph{complete} map of the environment. Though point cloud aggregation can build such a map, it faces challenges when the vehicle travels at high speeds~\citep{han2021planning}. Finally, since LiDAR emits lasers into the environment, dust and snow can interfere with the measurement, and outside observers can detect the vehicle from the emitted lasers.

Cameras, on the other hand, provide a number of benefits over LiDAR. Cameras provide high-resolution semantic and geometric information, stealth due to their passive sensing nature, are less affected by dust and snow, and are considerably cheaper. Hence, a camera-only off-road terrain perception system can potentially reduce the hardware cost, improve the system robustness at high speeds, and open up new possibilities for off-road navigation under extreme weather conditions and where stealth is desired.

Perhaps unsurprisingly, similar motivations have spurred recent major efforts of camera-only perception for \emph{on-road} navigation~\cite{chitta2021neat,hendy2020fishing,li2022bevformer,philion2020lift,yang2021projecting}. This task mainly focuses on Bird's Eye View~(BEV) semantic segmentation to assess traffic conditions. One notable work is Lift-Splat-Shoot (LSS)~\cite{philion2020lift}. The core of LSS consists of a ``lift" operation that predicts a categorical distribution over depth for each pixel and a ``splat'' operation to fuse the image features and project them to the BEV space. %We thus wonder, can we apply the same approach for off-road terrain perception?
%LSS does not directly supervise the intermediate lift and splat steps, instead the model is trained end-to-end where the loss function is defined on the final semantic BEV map prediction and training the model requires a large semantically labeled dataset.
LSS and related work are entirely data-driven, so they can predict complete maps and are more robust to sensor noise and projection errors. But their applicability to off-road perception is challenged by several barriers. First and foremost, they only predict a ground semantic BEV map without any 3D terrain information that is critical for planning and control in off-road environments. %Predicting such features requires accurate 3D depth estimation, but since the depth prediction in LSS is not directly supervised, it could fail to predict the terrain geometry accurately.
Second, they are usually not optimized for real-time operation. For example, LSS predicts depth as a categorical distribution along the camera frustums to enable end-to-end learning, but this comes at a price: the size of the frustum features is large, creating a time and memory bottleneck, especially for field robots with limited hardware capabilities. Finally, to train such models, we need large-scale labeled terrain datasets. But, to the best of our knowledge, there are no such datasets for complex off-road terrains yet.

To this end, we design and implement \bevnet{}, a real-time, camera-only terrain perception system that enables high-speed driving of a passenger-scale Polaris~\cite{rzr} vehicle on complex off-road terrains. We make several design choices and innovations to make \bevnet{} suitable for off-road perception. First, \bevnet{} supports multi-view RGB with \emph{optional stereo depth} as inputs. Using stereo depth provides valuable geometric context that greatly improves prediction accuracy. Second, we enhance the predicted depth via an auxiliary loss on the output depth values. This extra supervision teaches the model to \emph{correct} and \emph{complete} the (potentially inaccurate) input stereo depth. This turns out to be critical for accurately estimating terrain geometry. To create ground-truth depth images, we build a complete map of the environment offline by aggregating LiDAR scans and removing outliers from the entire point cloud. Note that LiDAR is only used to create the training dataset and is not needed in deployment. Third, we make \bevnet{} more than $5\times$ faster than LSS by lifting each image feature into a \emph{single} 3D point and use a soft quantization technique in the ``splat'' step to keep the model end-to-end trainable. Lastly, \bevnet{} predicts a \emph{multi-layer} BEV map that captures both ground and overhanging terrain features.
%images as input and lifts the features into a pseudo-pointcloud~\cite{qian2020end} using predictions of a pixel-wise depth prediction network. Then, the points are projected into a coherent BEV feature map, which is finally decoded into a multi-layer terrain map. Each layer of the terrain map captures a crucial aspect of the semantic or geometric properties of the terrain. The overall architecture is shown in Figure~\ref{fig:network}. To improve the speed and accuracy, our model is different from LSS in several ways: 1) we propose to supervise the pixel-level depth prediction network to enhance the estimated depth which is critical for capturing terrain geometry. 2) To maintain the elevation information in the projection step, we employ a neural network to embed the point elevation values into the BEV feature map. 3) We estimate a single point depth instead of a distribution, making the pointcloud size significantly smaller. To keep \bevnet{} end-to-end trainable using a smooth quantization~\cite{qian2020end} in the projection step.

\bevnet{} is the first off-road, camera-only perception system for joint BEV semantic and geometric terrain mapping in a unified feed-forward model. To train and evaluate our model, we collect a new challenging large-scale, off-road dataset from different environments consisting of both \emph{on-trail} and \emph{off-trail} driving scenarios with \emph{extreme elevation changes}. We believe our dataset better covers the diversity of the off-road driving challenges compared to RELLIS-3D~\cite{jiang2021rellis}, a publicly available dataset which is captured from a single environment and mainly consists of on-trail driving scenarios with limited elevation changes. We show that \bevnet{} outperforms recent baselines in semantic and elevation estimation, while being much faster. Finally, we deploy \bevnet{} inside a full navigation stack to have a Polaris vehicle traverse a 1.1~kilometer route over snow-covered hills. %\nolan{'til the landslide brought the vehicle down}

\section{Related Work}
\textbf{On-road BEV perception.} LiDAR and cameras are commonly used sensors in on-road autonomous driving perception systems~\cite{lang2019pointpillars,philion2020lift,zhu2020cylindrical}, providing crucial information about surrounding objects and their semantics. Convolutional Neural Networks (CNNs) have shown exceptional performance in image~\cite{chen2022vision} and point cloud~\cite{zhu2020cylindrical} segmentation, and they have become the core of perception systems in on-road scene understanding. Although many of these systems rely on LiDAR ~\cite{lang2019pointpillars,sadat2020perceive}, there has been increasing interest in using cameras due to their lower cost. In camera-based methods, a critical aspect is learning to project pixel-wise features to the BEV space. Lift-Splat-Shoot~\cite{philion2020lift} adopts a \emph{backward} projection scheme that lifts the image features using predicted depth and then splats the features into BEV space. SimpleBEV~\cite{harley2023simple} and BEVFormer~\cite{li2022bevformer} perform \emph{forward} projection from a set of grid points in the BEV space to retrieve their corresponding image features. Another line of work~\cite{chitta2021neat,roddick2020predicting,saha2022translating,zhou2022cross} learns the projection with an attention mechanism.  \bevnet{} adopts a backward projection scheme since it makes full use of image pixels and does not assume a flat ground. Moreover, \bevnet{} leverages stereo depth completion and soft-quantization for projection. This improves both the speed and accuracy of \bevnet{}.

\textbf{Off-road terrain modeling.} Off-road terrains often exhibit large variations in ground elevations which can significantly affect terrain traversability. Hence, there has been a plethora of work on geometric terrain mapping. A frequently used representation is a 2.5D elevation map by aggregating point measurements from LiDARs or stereo cameras~\cite{fankhauser2018probabilistic,forkel2022dynamic,forkel2021probabilistic,miki2022elevation,stolzle2022reconstructing,triebel2006multi}. In more complex environments where overhanging objects need to be considered, a voxel-based representation~\cite{bajracharya2013high} or a multi-level surface map~\cite{triebel2006multi} are more effective in capturing  detailed geometric information. In practice, obstacles or terrain discontinuities leave areas with missing values in the elevation map, leading to suboptimal motion planning. Inspired by data-driven image in-painting methods, recent works~\cite{vsalansky2021pose,stolzle2022reconstructing} propose self-supervised learning to reconstruct the occluded area from an incomplete elevation map.

Besides geometric terrain modeling, semantics also play a critical role in traversability assessment~\cite{bajracharya2009autonomous,kelly2006toward,manduchi2005obstacle,  maturana2018real}. For instance, tall grass appears as obstacles yet is traversable, whereas large puddles are perceived as a flat surface but can be dangerous. Semantic segmentation is typically done in the image space ~\cite{bajracharya2009autonomous,  schilling2017geometric,suleymanov2016path} or the BEV space~\cite{maturana2018real,shaban2022semantic}. Since planning is more convenient in the BEV space, it is a common practice to project the pixel-wise segmentation into a BEV cost map using LiDAR or stereo cameras~\cite{bajracharya2009autonomous, maturana2018real}. More recently, \citet{shaban2022semantic} present an end-to-end trainable, recurrent CNN that builds a temporally consistent BEV semantic map from LiDAR. For a comprehensive literature review in traversability estimation for mobile robots, we refer readers to surveys~\cite{borges2022survey,sevastopoulos2022survey}.

Existing off-road terrain perception approaches are usually designed for low-speed operations. Thus, they are able to gather dense sensor measurements to filter out the sensor noise to estimate a good terrain model. Some systems support high-speed operations but they are limited to on-road, flat terrains. In contrast, \bevnet{} is designed for \emph{high-speed} operations on \emph{any terrain}. 
It achieves low latency by using a compact, multi-layer terrain  representation ~\cite{triebel2006multi}, while at the same time maintaining accuracy at high speed with end-to-end BEV perception~\cite{philion2020lift, shaban2022semantic}. Unlike existing systems that require complex algorithms and careful tuning to maintain real-time operations and consistent mapping, \bevnet{} is a simple feed-forward neural network. It is also fast and can be improved as more training data is available.

\textbf{Planning.} To navigate a vehicle in off-road environments, a motion planner can leverage the perceived terrain features to plan safe and efficient trajectories. The terrain features are usually converted into costs for a planner to rank and assess the risk of trajectories~\cite{cai2022risk,cai2022probabilistic,dashora2022hybrid,fan2021learning}. In our experiments, we illustrate how to use the MPPI planner~\cite{williams2017information} for motion planning with terrain features and robot capability considered.

\section{Off-Road Terrain Modeling}
\label{sec:terrain-modeling}

\begin{figure*}[t]
    \centering
    \includegraphics[width=1\textwidth]{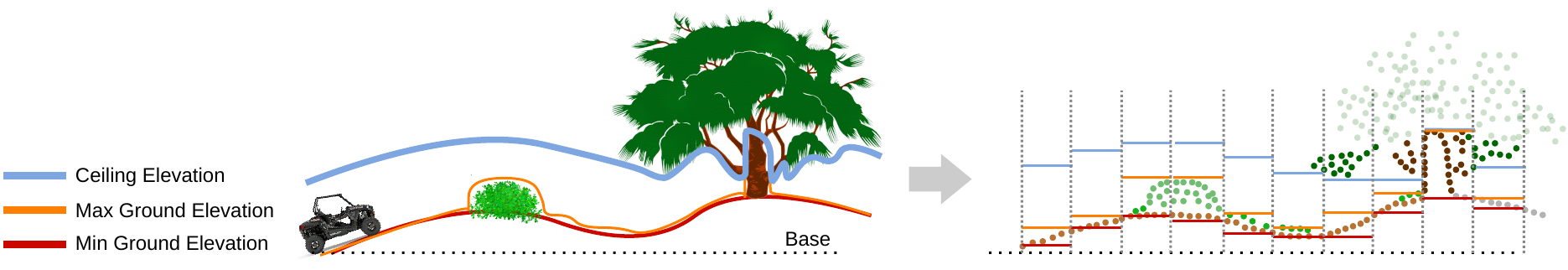}
    \caption{{\bf Left} Multi-layer terrain representation. The ground layer consists of the minimum and maximum ground elevations to capture the sizes of porous objects like bushes and also preserve sharp elevation changes after discretization. The ceiling layer represents the geometry of the overhanging objects (e.g., canopy) that might potentially block the path. {\bf Right} In our practical implementation, we obtain a dense point cloud of the environment, discretize the $x$-$y$ plane into 2D grid cells (shown in 1D for simpler visualization), and compute the elevation values by analyzing the point distribution in each cell.  We ignore anything higher than the desired vertical clearance. For each layer and each cell, we count the number of semantic points for each class and compute a normalized histogram as the semantics of that cell.}
    \vspace{-6mm}
    \label{fig:terrain-model}
\end{figure*}

To enable a robot to drive safely and efficiently on off-road terrains, it is crucial to understand the traversability of its surroundings. Terrain traversability is the amount of cost or effort to traverse over a specific landscape. While many factors affect terrain traversability, we consider three primary factors: \textit{semantics}, \textit{geometry}, and \textit{robot capability}.

\textbf{Semantics.} The semantics of terrain refers to the classes of objects (e.g., bush, rock, tree) or materials (e.g., dirt, sand, snow) occupying the terrain. Different semantic classes typically have different physical properties, such as friction and hardness, which can affect the capabilities of the vehicle.  For example, since dirt can supply more friction than snow, a vehicle can drive faster on a dirt road than on snowy ground. Moreover, off-road vehicles have higher chassis and better suspension, so they can traverse over bushes and small rocks, albeit at lower speeds due to the increased resistance and bumpiness. Hence, the semantics of terrain encodes a rich spectrum of traversability.

\textbf{Geometry.} Off-road terrains are typically non-flat. A vehicle may not have enough power to climb a steep slope, and driving along a slide slope at high speed poses a significant risk of rolling over. Additionally, the geometry of objects also affects terrain traversability. For instance, a large bush is harder to traverse than a small bush. Hence, understanding the geometry of terrain is another important aspect of traversability assessment.

\textbf{Robot capability.} A vehicle's physical and mechanical properties play another important role in terrain traversability. A bigger and more powerful vehicle can traverse over larger bushes or rocks than a smaller vehicle with less power. Since robot capability is an intrinsic property of the robot and is independent of terrain properties, we consider robot capability when designing the cost function later in Section~\ref{sec:planning}. To some extent, robot capability is also considered when generating the training dataset, as described in Section~\ref{sec:dataset}.

\subsection{Multi-layer Terrain Representation}
Since a ground vehicle traverses a 2D surface, it is convenient to use a gravity-aligned, 2D top-down grid map~\cite{elfes1989using} to represent the terrain. Here we consider \emph{local navigation}, where the map provides the vehicle with instantaneous information about its surroundings. Hence, we fix the size of the map and let the map ``move'' with the vehicle so that the vehicle stays at the center. This is commonly referred to as the \emph{local map}. We do not consider building a global map in this work, though if required, we can leverage existing global SLAM algorithms to stitch the local maps together to obtain a global map.

The key question is what kind of terrain features to store in the top-down map. Previous work usually stores semantic classes~\cite{maturana2018real,shaban2022semantic} or elevations~\cite{fankhauser2018probabilistic,stolzle2022reconstructing} for each grid cell. These representations have two key drawbacks. First, they do not model the hardness or porousness of the terrain, and hence they cannot capture the difference in traversability between a small and large bush. Second, they either do not consider overhanging objects or merge the semantic information of overhanging objects with ground objects. This would result in an inaccurate terrain model because the effect on traversability from overhanging objects is different from ground objects due to the geometry and the lack of gravity-induced force. 

To address these issues, we extend the idea of MLS map~\cite{triebel2006multi} and propose a multi-layer terrain representation illustrated in Figure~\ref{fig:terrain-model}. It consists of two layers, a \textbf{ground} layer that captures terrain properties on the ground and a \textbf{ceiling} layer that models overhanging objects. For each layer, we model their semantic and geometric properties separately as follows:

\textbf{Ground layer.} For each map cell on the ground, we store the semantic probability distribution $\mathcal{C}_\text{ground}$ and elevation statistics $\mathcal{H}_\text{ground}$ of the terrain. The categorical distribution ${\mathcal{C}_\text{ground} \in \mathbb{R}^K}$ stores the relative proportions of the $K$ semantic classes occupying each cell. We use the full distribution of semantic classes instead of a single class label to reduce the information loss caused by discretization (e.g., a map cell may contain both ``dirt'' and ``bush'' if it is at the boundary between dirt and a bush). The elevation statistics $\mathcal{H}_\text{ground}$ contains the minimum and maximum elevation values $h_\text{min}$ and $ h_\text{max}$ on the ground. This allows the height of porous objects (such as grass and bushes) to be captured separately and resolves sharp elevation changes due to objects such as rocks and trees. %\amir{Can not see how it resolves sharp elevation changes?}

\textbf{Ceiling layer.} The ceiling layer models \emph{overhanging} objects, such as canopies and tree branches. The semantic information $\mathcal{C}_\text{ceiling}$ is similar to $\mathcal{C}_\text{ground}$ but stores the semantic distribution of overhanging objects. The elevation information $\mathcal{H}_\text{ceiling}$ stores the height of the lowest overhanging point $h_\text{ceiling}$. If no overhanging points are present (e.g., on open terrains), we set $h_\text{ceiling} = h_\text{min} + h_\text{clearance}$, where $h_\text{clearance}$ is a predefined constant of the desired vertical clearance.

This two-layer terrain representation captures a number of properties that are crucial for off-road navigation: the separate modeling of ground and overhanging semantics allows a robot to reason about semantic traversability more accurately, and the ground elevation statistics captures the hardness and sizes of ground objects (refer to Figure~\ref{fig:terrain-model}). These features can then be input to a cost function to build an accurate traversability cost map for planning (see Sec.~\ref{sec:planning}).

\subsection{Computing the Terrain Representation}
\label{sec:elevation-computation}
This section illustrates a practical approach that leverages LiDAR sensors to build the proposed multi-layer terrain representation. This process is summarized in Figure~\ref{fig:terrain-model}. While our goal is real-time visual terrain modeling, we can benefit from accurate geometric sensors and extensive offline processing to produce high-quality ground-truth data for training that is otherwise challenging to obtain in real time. 

\textbf{Point cloud aggregation.} Given recorded LiDAR scans, we use offline SLAM tools~\cite{cartographer} to generate gravity-aligned poses for each scan and then aggregate them to obtain a globally aligned dense point cloud. We clean up the point cloud by removing points that are potentially outliers (see the supplementary for more details). Since this is done offline, we can thoroughly analyze the point cloud without worrying about the computational cost. Note that we could also use the stereo cameras for creating the dense point cloud, but we have found that stereo cameras have a shorter range and produce spurious depths on porous objects and the ground.

\textbf{Divide and sort.} We divide the $x$-$y$ plane into 2D grid cells with a desired resolution and sort points that fall into a cell $C(i, j)$ by their $z$-coordinates in ascending order, giving us a sorted point set $\{(x_k, y_k, z_k)\in C(i,j)\}$ for each cell.

\textbf{Compute elevations.} 
We iterate over the points in each cell, starting from the lowest elevation. The minimum ground elevation of a cell is computed as
\begin{equation}
h_\text{min} = \cfrac{1}{m}\sum_{k=1}^m z_k,
\end{equation} where $m$ is a tuning parameter to smooth out the sensor noise. Cells without enough points are marked as unlabeled. Then, we compute the $z$-gaps of consecutive points as ${\Delta z_i = z_{i+1} - z_{i}}$. Letting $z_\text{gap}$ denote a predefined constant corresponding to the minimum gap between the ground and the ceiling, if ${\Delta z_i > z_\text{gap}}$, then $z_{i+1}$ is considered the lowest overhanging point, and we set:
\begin{align}
h_\text{max} &= z_{i} \\
h_\text{ceiling} &= z_{i+1}.
\end{align}
If no gap is found, then $h_\text{max}$ is set to the highest point in the cell, and $h_\text{ceiling} = h_\text{min} + h_\text{clearance}$ where $h_\text{clearance}$ is the vehicle's vertical clearance. 

\textbf{Compute semantics.}
The ground and ceiling layers have separate semantic maps. Given a labeled point cloud (Section~\ref{sec:dataset}), we assign each point to either the ground layer or the ceiling layer. Specifically, for a point $(x, y, z)$,
\begin{equation}
(x,y,z) \in  \begin{cases}
    \text{ground}, & \text{if $z \leq h_\text{max}$}.\\
    \text{ceiling}, & \text{if $h_\text{max} < z < h_\text{ceiling}$}.\\
    \text{ignored}, & \text{otherwise}.
  \end{cases}
\end{equation}
After the partitioning, we compute a normalized histogram for each cell 
\begin{equation}
[p_1, p_2, \dots, p_K] = \cfrac{1}{\sum_{k=1}^K n_k} \; [n_1, n_2, \dots, n_K],
\end{equation}
where $n_k$ is the number of points of class $k$ in that cell.

\section{\bevnet{}}

In the previous section, we introduced our terrain representation. When a vehicle is deployed in a new off-road environment, it has no knowledge of the terrain and thus must build the terrain representation from its onboard sensors in real time. In this section, we introduce our terrain inference engine \emph{\bevnet{}}, which predicts the terrain representation from its onboard RGB or RGB-D cameras. 

\subsection{Overview}

\begin{figure*}[t]
    \centering
    \includegraphics[width=1\textwidth]{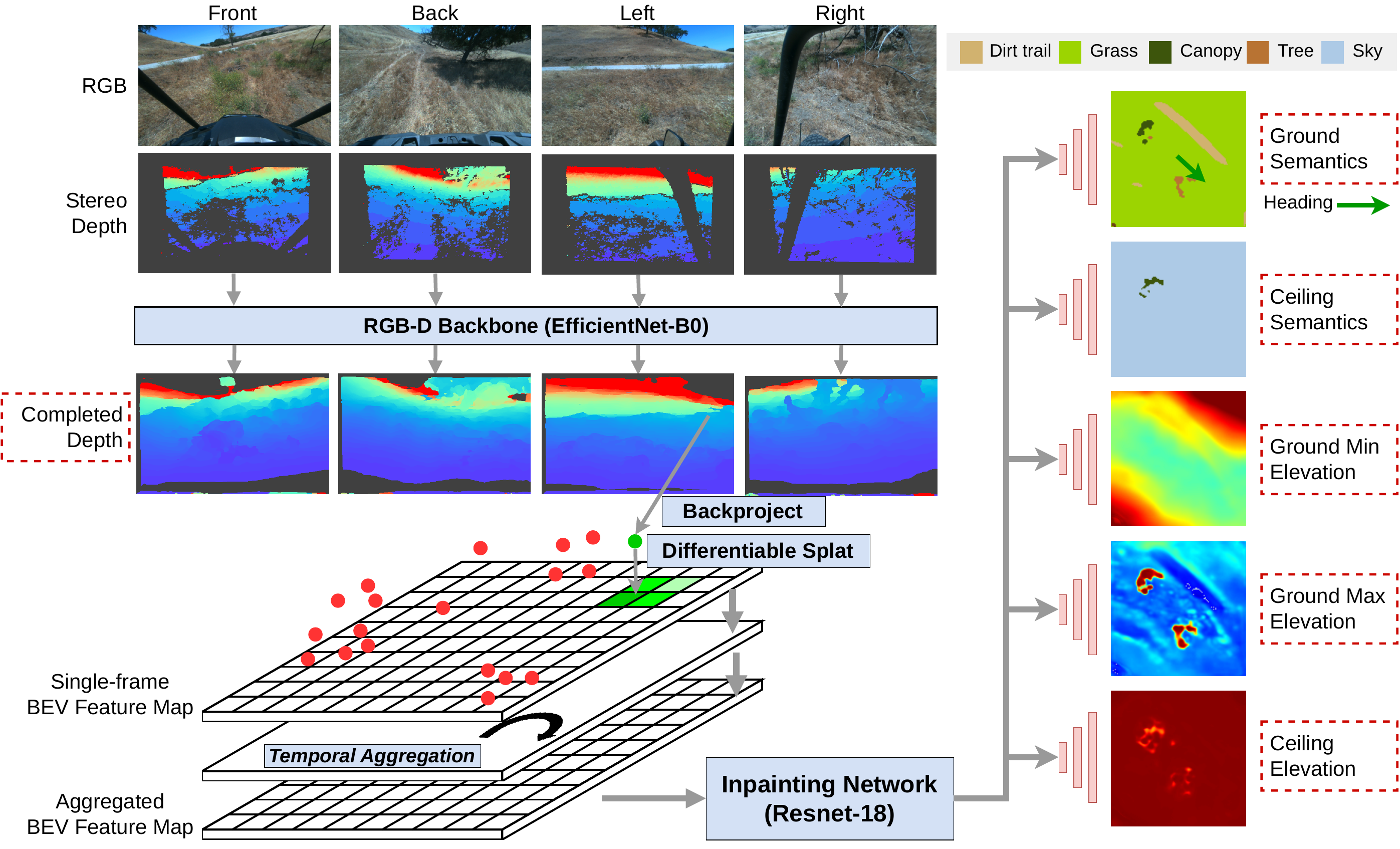}
    \caption{\textbf{Architecture of \bevnet{}.} RGB images with stereo depths are input to the RGB-D backbone to obtain completed depths with semantic features. Note that completed depths exclude pixels (colored in dark gray) that are potentially self-hits or sky. The per-pixel image features are back-projected to 3D to compute their terrain embeddings. The terrain embeddings are then splatted onto the BEV feature map with soft quantization. The feature map is temporally aggregated and finally passed to the multi-head inpainting network to predict a multi-layer terrain map. Red dotted boxes indicate where losses are applied during training.
    }
    \label{fig:network}
    \figurespace
\end{figure*}
\bevnet{} is a single neural network that takes RGB or RGB-D images and jointly predicts terrain semantics and geometry. Figure~\ref{fig:network} illustrates the overall pipeline. First, multiple RGB or RGB-D sensors are passed into the RGB-D backbone. The backbone produces two outputs for each camera: a dense and corrected depth image, and a 2D feature map. Each valid~(i.e., neither self-hit nor sky) pixel in the depth image is back-projected to a 3D point using the corresponding camera intrinsics and extrinsics. We compute the terrain embedding for each 3D point that encodes semantic and geometric features. The per-point terrain embeddings are then down-projected and aggregated in the gravity-aligned BEV space with a fast, differentiable splat operation. The resulting BEV feature map is decoded into multiple semantic and elevation maps via an inpainting network. These maps are converted into a costmap by mapping the semantic categories and geometric features via a cost function that accounts for robot capability. The costmap can then be used by a local planner to find the best trajectory to reach a waypoint. In the following sections, we describe each component in detail.

\subsection{Depth Completion and Correction}
To build an accurate terrain representation, the most crucial aspect is to have a good spatial understanding of the environment. Modern stereo cameras can provide dense and accurate depth at close range. However, the error in depth estimation increases quadratically over distance, and stereo matching works poorly for porous structures such as vegetation. Moreover, high-end stereo cameras~\cite{multisense} are equipped with high-sensitivity, low-distortion monochrome sensors for stereo matching, but they are not engineered for wide field-of-view semantic perception like RGB cameras. In Figure~\ref{fig:network}, we show the stereo depth output with the wide-angle RGB image on our hardware platform. The stereo depth only occupies a central area of the image, and there are many missing depth pixels due to failures in stereo matching. On the other hand, while there have been substantial advances in monocular depth estimation, they have only been shown to work well in structured environments where prior knowledge of object sizes and depth can be learned. Off-road terrains are often much less structured, and we find that solely using RGB information is inferior compared to a stereo system. To combine the rich semantic information of wide-angle RGB cameras and the incomplete stereo depth, we perform \emph{Depth Completion}, which fills in missing depth pixels and corrects the errors in stereo depth in a data-driven fashion. 

Given an RGB image and a stereo depth image (note that these may come from two different cameras), we first transform the stereo depth points into the RGB image to obtain an aligned RGB-D image $I$. Then, we pass the RGB-D image into a U-Net similar to \citet{jaritz2018sparse} to get a dense depth map. The depth map does not have to be full-resolution, and we find $8\times$ downsampling provides a good trade-off between speed and accuracy. We adopt a classification approach for depth completion because it captures depth discontinuity better than regression~\cite{reading2021categorical}.

Given the completed depth image and the feature map, we use the camera intrinsics $\mathbf{K}$ and extrinsics $[\mathbf{R}, \mathbf{t}]$ to compute the 3D location of each pixel in the gravity-aligned BEV space. Specifically, given a 2D pixel $(u, v)$ with depth $d$, we transform it to the vehicle-centered, gravity-aligned BEV frame:
\begin{equation}
    [x, y, z]^\top = \mathbf{R}\mathbf{K}^{-1}[u,v,d]^\top + \mathbf{t} .
\end{equation}

\textbf{Terrain embedding.} For each 3D point, we use the corresponding image embedding as the semantic embedding~$f_\text{sem}$. For elevation, we apply a multi-layer perceptron (MLP) on $z$ to obtain the elevation embedding: $f_\text{elev} = \mathrm{MLP}(z)$. We concatenate the semantic and elevation embeddings together and apply another MLP to obtain the per-point terrain embedding: $f = \mathrm{MLP}(\mathrm{concat}(f_\text{sem}, f_\text{elev}))$.

\subsection{Fast and Differentiable BEV Projection}

Given the 3D point set $\{(x, y, z, f)\}$, the next step is to project the points onto the BEV map by rounding each point into the nearest integer grid coordinates  $(\mathrm{round}(x/r), \mathrm{round}(y/r))$, where $r$ is the side length of each map cell. The main disadvantage of hard quantization is that we can no longer correct the grid coordinates via back propagation due to the loss of gradient w.r.t grid coordinates. Inspired by related work in differentiable geometric learning~\cite{qian2020end}, we adopt a \textbf{soft quantization} approach, where we ``splat'' each point feature into the 4 neighboring map cells with weights computed by bilinear interpolation. This allows the depth completion module to learn to adjust its depth prediction by minimizing the loss in the projected map.

\textbf{Local feature aggregation.} Since multiple points may fall into the same map cell,
%we compute the cell-level terrain embedding by taking the weighted average of the per-point embedding $f_\text{cell}(i, j) = \sum w_k f_k / \sum w_k$ where $w_k$ are the weights computed by soft quantization.
we compute the weighted average of the embeddings in each cell, where the weights are given by the soft quantization to create a grid-based BEV feature map. Let $S(i,j)=\{(f_1, w_1), \dots, (f_N, w_N)\}$ denote the set of point embeddings, and their corresponding weights within cell $(i,j)$, the feature embedding for the cell is computed as
\begin{equation}
f_\text{cell}(i, j) = \frac{1}{W} \sum_{k=1}^N w_k f_k,
\end{equation}
where $W = \sum_{k=1}^N w_k$.

\textbf{Temporal feature aggregation.} Due to occlusion and errors in estimated depth, the map built from a single frame may not be sufficiently stable and complete. Hence we introduce an optional \emph{Temporal Aggregation} (TA) layer that aggregates BEV feature maps over time. The TA layer is a single ConvGRU similar to that of \citet{shaban2022semantic} but with one major difference: we use an orientation-stable odometry frame for aggregation instead of an ego-centric frame to better preserve fine-grained details. This is similar to the classical sliding window approach~\cite{maturana2018real} where we shift the map and integrate the current sensor measurements,  but we perform the shifting and aggregation on the feature map with a recurrent network.

\subsection{Multi-head Terrain Inpainting Module}
The terrain inpainting module decodes the BEV feature map into complete semantic and elevation maps. For semantic maps, it predicts a $H\times W \times K$ tensor to represent the probability distribution of the semantic classes, where $H, W$ are the height and width of the map. For elevation maps, it predicts a $H\times W\times 1$ tensor to represent the corresponding elevation values. We adopt a U-Net as the inpainting module with a shared encoder and multiple convolutional decoding heads. Each head predicts a specific type of terrain map. This works better than a single decoding head with multiple channels because the semantic and elevation maps have different output spaces.

\begin{figure*}[t]
    \centering
    \includegraphics[width=1\textwidth]{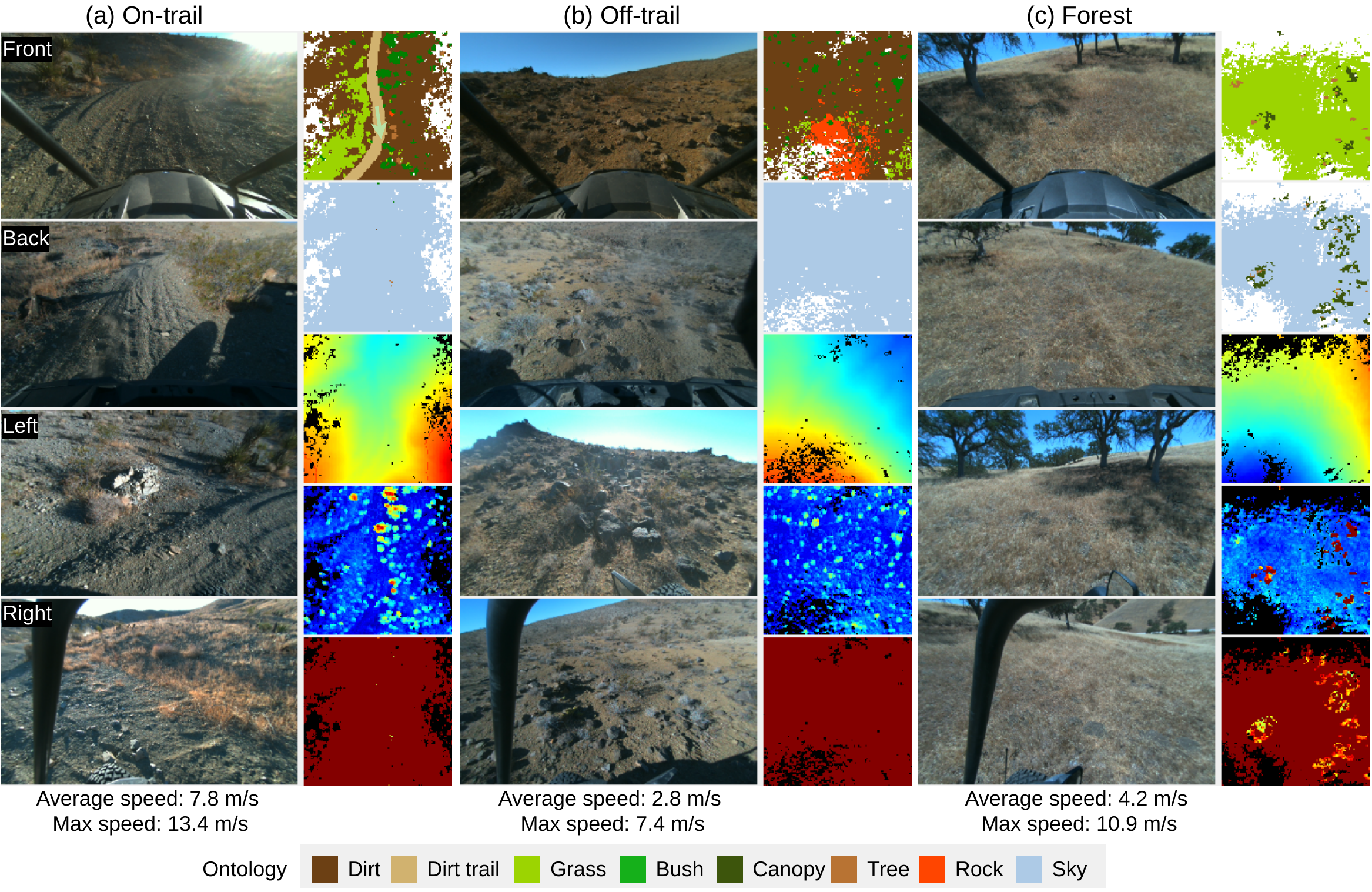}
    \caption{Our dataset consists of diverse off-road environments. \textbf{a)} vehicle driving at high-speed on a trail with bushes, grass, rocks, and Joshua trees on the two sides; \textbf{b)} vehicle driving off-trail on a hilly terrain with scattered grass, bushes, rocks and Joshua trees; \textbf{c)} vehicle driving on a hilly terrain with tall trees and overhanging canopy. For each environment, we also show (from top to bottom) the ground semantics, ceiling semantics, min ground elevation, max ground elevation, and ceiling elevation. For the ceiling semantic map, we use the \emph{sky} class to mark areas where no overhanging objects are present. The max ground and ceiling elevation maps are \emph{offsets} to the ground elevation.}
    \label{fig:ontology}
    \figurespace
\end{figure*}

\section{Implementation Details}
\subsection{Hardware Platform}

We collect our training and test data on a modified Polaris RZR vehicle~\cite{rzr} shown in Figure~\ref{fig:overview}, which is capable of driving on off-road terrains at speeds up to $20~\mathrm{m/s}~(45~\mathrm{mph})$. The vehicle is equipped with 4 MultiSense stereo cameras~\cite{multisense} and 3 Velodyne 32-beam LiDAR sensors. LiDAR is only used for generating ground-truth with the method discussed in Section~\ref{sec:elevation-computation} and is not used during testing.

\subsection{Dataset}
\label{sec:dataset}

We collected our training and validation data on the actual vehicle in several off-road environments with diverse appearances (Figure~\ref{fig:ontology}). We collected 5 sequences of manual driving data, totaling 20k frames. For each sequence, we ran Google Cartographer~\cite{cartographer} to obtain gravity-aligned robot poses for building the ground-truth terrain maps and depth maps.

\textbf{Create elevation maps.} We follow the algorithm in Section~\ref{sec:elevation-computation} to generate the minimum ground, maximum ground, and ceiling elevation maps. For each time step $t$, we aggregate 300 LiDAR scans (running at $10~\mathrm{Hz}$) with a time range of $(t - 150, t + 149)$. Note that the vehicle is always at the origin. We set $m=3$, $z_\text{gap}=1~\mathrm{m}$, and $h_\text{clearance}=3~\mathrm{m}$.

\textbf{Create semantic maps.} We ask human labelers to annotate 6k LiDAR scans. These scans are carefully selected such that they contained the relevant semantic objects. These annotated LiDAR scans are aggregated and projected to the BEV ground map or the BEV ceiling map using the algorithm in Section~\ref{sec:elevation-computation}.  We identify 7 semantic categories: \emph{dirt}, \emph{dirt-trail}, \emph{grass}, \emph{bush}, \emph{canopy}, \emph{tree}, and \emph{rock}. It is straightforward to extend the ontology and finetune the model if additional data is available, as we show in Section~\ref{sec:planning}.

\textbf{Create dense depth maps.} We aggregate 50 LiDAR scans for each time step and project the point cloud to each RGB camera. For each pixel, we keep the smallest depth value.

\subsection{Pseudo Labeling}
To leverage unlabeled data, we train a LiDAR segmentation network to predict \emph{pseudo labels}~\cite{lee2013pseudo} for the unlabeled points in the training set. Specifically, we train Cylinder3D~\cite{zhu2020cylindrical} on the labeled LiDAR points and predict the labels for all the unlabeled LiDAR points. We use the pseudo-labeled LiDAR points to build our ground-truth semantic maps for training. We \emph{do not} apply pseudo labeling on the validation set. In Section~\ref{sec:exp-pseudolabel} we evaluate the effect of pseudo labeling.

\subsection{Training}
We split the dataset into 15k frames for training and 5k frames for validation by cutting each sequence into two non-overlapping train and validation segments. We manually labeled 4500 training frames and 1500 validation frames, and we pseudo-labeled the remaining training frames. All frames have elevation labels. The images are rectified and resized to $512\times 320$. We crop and warp the images such that they all have the same intrinsic parameters. We train \bevnet{} on 4 Nvidia A40 GPUs with a batch size of 8 and the Adam optimizer~\cite{kingma2015adam}. All models (including applicable baselines) are trained for 100k iterations. The map size is $51.2~\mathrm{m} \times 51.2~\mathrm{m}$ with a resolution of $0.4~\mathrm{m}$. There is no limit on the elevation range.

\textbf{Loss.} We first pre-train the RGB-D backbone on the depth completion task. We discretize the depth into 128 bins with a max range of $25.6~\mathrm{m}$ and apply a cross-entropy loss. Then, we train the whole network end-to-end. For the semantic maps, we use an unweighted cross-entropy loss. For the elevation maps, we use a Smooth-L1 loss~\cite{smoothl1} with transition parameter $\beta=0.2$. Instead of predicting $h_\text{max}$ and $h_\text{ceiling}$ directly, we predict their \emph{offsets} to $h_\text{min}$. The total loss is $L = L_\text{semantic} + 0.1 L_\text{elevation} + 0.1 L_\text{depth}$.

\begin{table*}[t]
  \centering    
  \begin{tabular}{cl|llllllll|ccc|l}
    \toprule
    & Method & Dirt & Dirt-trail & Grass & Bush & Canopy & Tree & Rock & mIoU $\uparrow$ & \multicolumn{3}{c|}{Elevation (MAE) $\downarrow$} & Time (ms) $\downarrow$\\
    \hline
    \parbox[c]{1mm}{\multirow{3}{*}{\rotatebox[origin=c]{90}{\textbf{RGB}}}} & LSS & 0.726 & 0.621 & 0.921 & 0.241 & 0.042 & 0.081 & 0.141 & 0.396 & 0.375 & 0.323 & 0.015 & 202\\
    % 85.05k
    &SimpleBEV & 0.727 & 0.614 & 0.922 & 0.245 & 0.040 & 0.041 & 0.106 & 0.385 & 0.417 & 0.322 & 0.018 & 110\\    
    & \bevnet{} & 0.730 & 0.617 & 0.919 & 0.257 & 0.071 & 0.081 & 0.177 & 0.407 & 0.361 & 0.315 & 0.018 & 25\\
    \hline
    % Effnet with separate RGB and depth blocks
    \parbox[c]{1mm}{\multirow{6}{*}{\rotatebox[origin=c]{90}{\textbf{RGB+Stereo}}}} & LSS & 0.762 & 0.656 & 0.926 & 0.353 & 0.096 & 0.141 & 0.235 & 0.453 & 0.257 & 0.286 & 0.016 & 207 \\
    & LSS one-hot depth & 0.751 & 0.642 & 0.919 & 0.308 & 0.094 & 0.098 & 0.202 & 0.431 & 1.028 & 0.379 & 0.019 & 28$^*$\\
    & SimpleBEV & 0.754 & 0.647 & 0.923 & 0.320 & 0.076 & 0.070 & 0.140 & 0.419 & 0.286 & 0.299 & 0.015 & 113\\
    & \bevnet{} & 0.765 & 0.666 & 0.926 & 0.380 & 0.125 & 0.145 & 0.274 & 0.469 & 0.244 & 0.277 & 0.015 & 28\\    
    & Seg \& Proj & 0.664 & 0.671 & 0.784 & 0.234 & 0.109 & 0.081 & 0.168 & 0.387 & 0.559 & 0.408 & 0.034 & - \\  
    & \bevnet{}-TA & 0.796 & 0.679 & 0.928 & 0.513 & 0.161 & 0.192 & 0.276 & 0.506 & 0.243 & 0.240 & 0.024 & 29\\
    \bottomrule
  \end{tabular}
  \caption{For semantics, we report the per-class IoU and mean IoU. For elevation, we report the Mean Absolute Error (MAE) for min. ground, max. ground, and ceiling elevation, respectively. We report the inference time on an RTX 3090. The inference time for LSS one-hot depth is an estimation. For Seg\&Proj we do not report its inference time since it depends on the implementation of temporal aggregation.}
  \label{tab:quantitative-pseudo}
  \figurespace
\end{table*}

\begin{figure*}[t]
    \centering
    \includegraphics[width=1\textwidth]{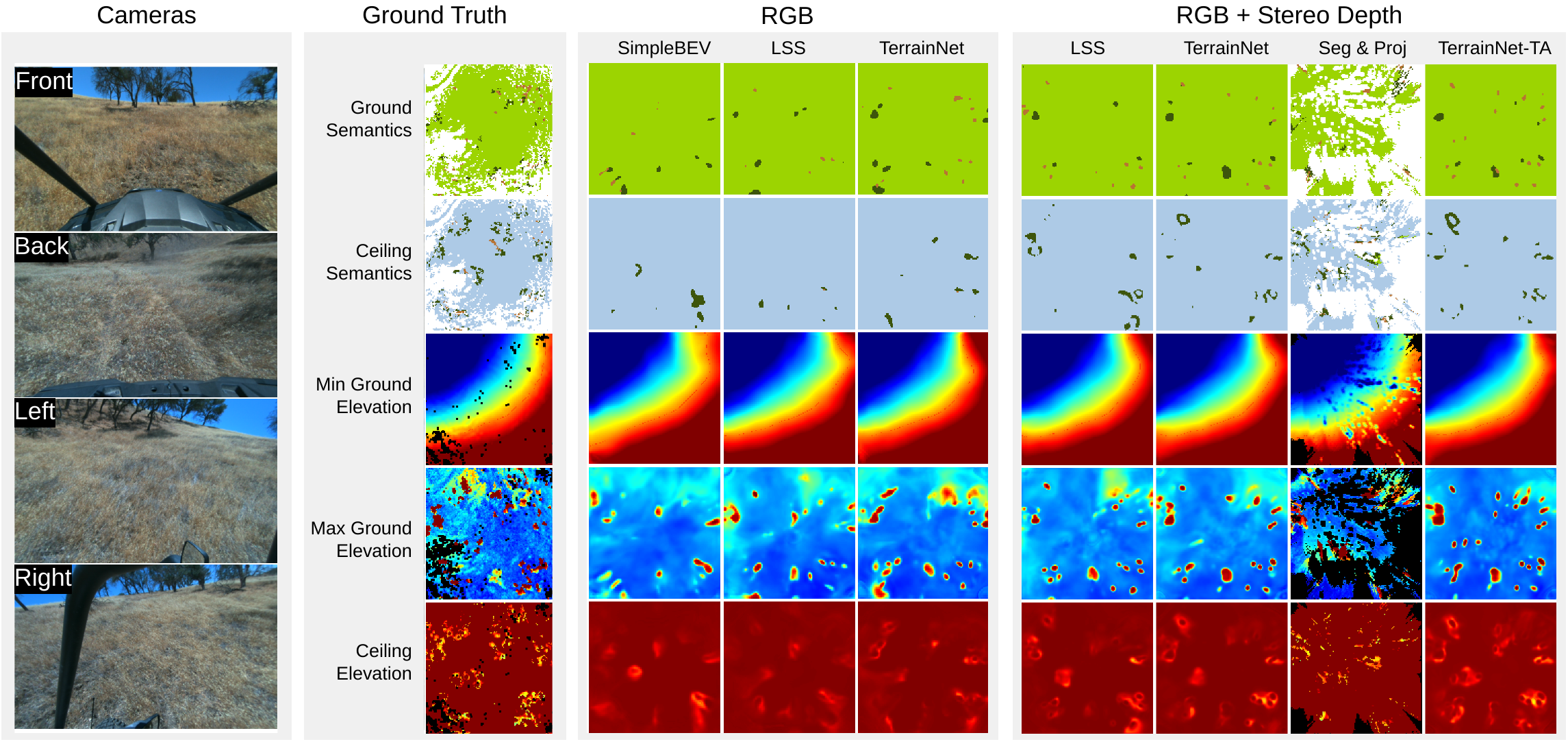}
    \caption{Qualitative comparison between \bevnet{} and the baselines in one of the forest environments (best viewed when zoomed in). In general, RGB-D models see more trees. Seg\&Proj shows many artifacts due to errors in stereo matching on the ground. \bevnet{}-TA shows the highest fidelity. See the supplementary material for more examples.}
    \label{fig:compare-baselines}
    \figurespace
\end{figure*}

\section{Experiments}

\subsection{Comparison with Existing BEV Perception Methods}
We quantitatively evaluate \bevnet{} in terms of its accuracy and speed.  We divide the experiments into a few sections, each focusing on a specific aspect. 

\subsubsection{Baselines}
There is a vast amount of literature on on-road BEV perception. Many of them use heavyweight neural nets, which are unsuitable for running on a compute-limited vehicle. To this end, we focus on three main BEV perception paradigms and pick the most representative baseline for comparison:

\textbf{Segmentation and Projection.}  We build a classical pipeline~\cite{guan2022tns,maturana2018real} where we first perform image segmentation and then project the pixel labels using the stereo depths. Since we only label the LiDAR points, we project the labeled LiDAR points to each image as the ground-truth segmentation mask. We train a state-of-the-art image segmentation network SegFormer-B1~\cite{xie2021segformer} on our dataset. Then, we use the same pipeline for generating the ground-truth maps to build a dense point cloud and project it down to the BEV map. We perform nearest neighbor inpainting to fill in the unknown space.

\textbf{Lift-Splat-Shoot.} LSS~\cite{philion2020lift} implicitly learns to predict a per-pixel depth map and splats the features along each depth ray, weighted by the depth distribution. Since vanilla LSS cannot predict elevation, we modified LSS to compute the terrain embedding in the same way as \bevnet{}. The main difference between LSS and \bevnet{} is that \bevnet{} has explicit depth supervision and does a ``one-hot'' splat of terrain embeddings onto the map.

\textbf{SimpleBEV.} SimpleBEV~\cite{harley2023simple} performs forward projection with a uniform splat of image features without considering per-pixel depth. While it has been shown to outperform LSS and many other strong methods for on-road driving, it is not clear if the same holds for \emph{off-road} driving where the terrain is not flat. We use a grid resolution of $128\times 128\times 32$ to cover a 3D volume of $51.2~\mathrm{m} \times 51.2~\mathrm{m} \times 51.2~\mathrm{m}$. (Using a larger $z$-resolution would make the memory consumption and training time prohibitive.)

We evaluate two versions of our system: the single-frame \bevnet{} and \bevnet{}-TA (i.e., with a temporal aggregation layer). For a fair comparison, LSS and SimpleBEV use the same image backbone and inpainting net as \bevnet{} with the same hyperparameters for training. We also train a RGB-only version of TerrainNet, LSS, and SimpleBEV by removing the input depth.

\subsubsection{Terrain Modeling Accuracy}

In Table~\ref{tab:quantitative-pseudo}, we compare \bevnet{} with the baselines in terms of their accuracy in modeling the terrain. For semantics, we use the standard IoU metric. For elevation, we use the per-pixel mean absolute error. 

\begin{figure}[t]
    \centering
    \includegraphics[width=0.95\columnwidth]{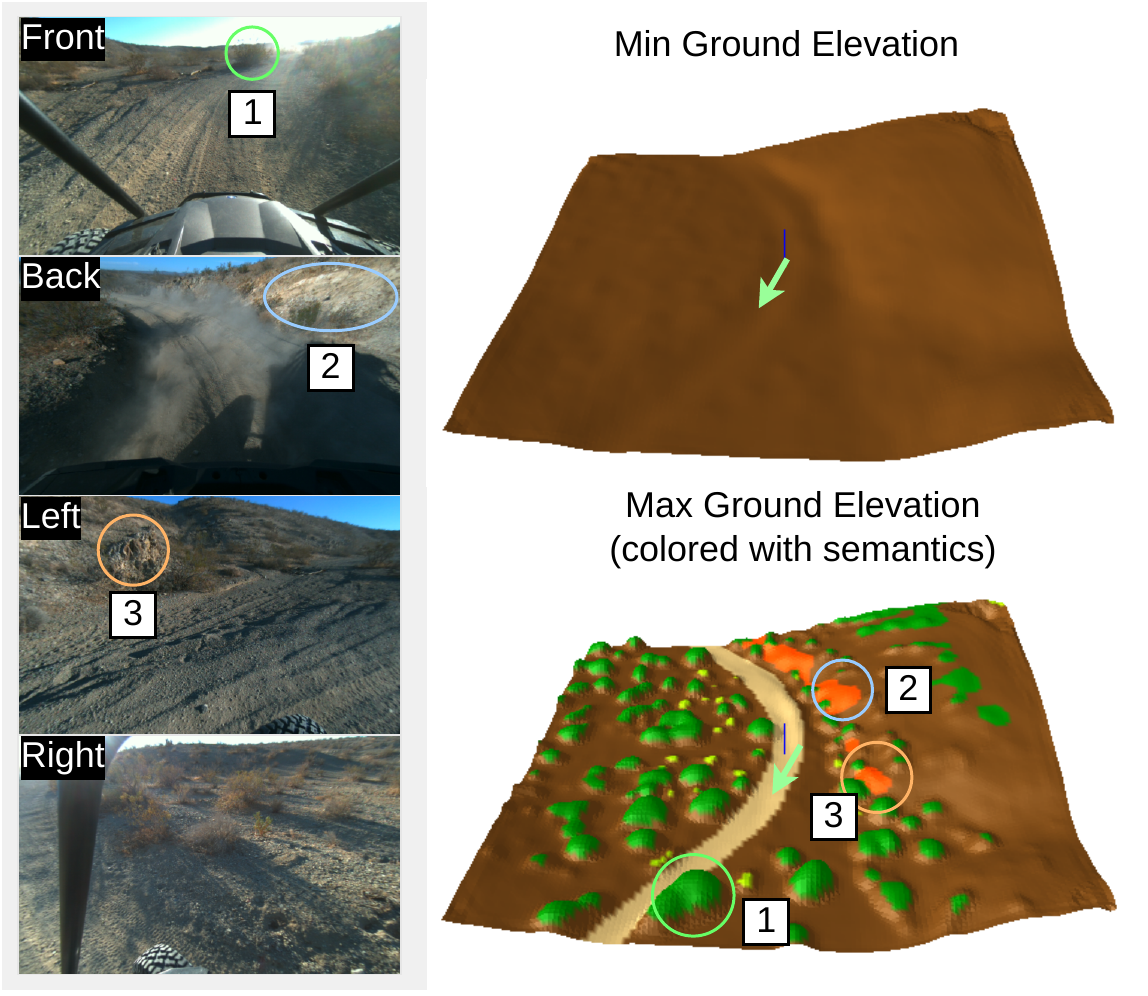}
    \caption{3D visualization of \bevnet{}-TA prediction on a high-speed on-trail sequence. The green arrow indicates the vehicle's heading. The Min Ground Elevation map shows a smooth ground support with porous objects such as vegetation removed. The Max Ground Elevation map contains the protruding vegetation on the ground. Note the bush far ahead (1) and rocks on the hills (2, 3) are well captured.}
    \label{fig:visualize-elevation}
    \figurespace
\end{figure}

\bevnet{} surpasses all the baselines for RGB and RGB-D inputs. The fact that \bevnet{} outperforms LSS shows that the explicit learning of per-pixel depth is beneficial. While it is possible to project a single depth for LSS during inference, the results are much worse, as shown in the \textit{LSS one-hot depth} baseline. SimpleBEV and Seg\&Proj perform worse than \bevnet{} and LSS. Figure~\ref{fig:compare-baselines} presents a qualitative comparison and Figure~\ref{fig:visualize-elevation} highlights a particular frame. The RGB-D models preserve more semantic details, with \bevnet{}-TA being closest to the ground truth. The artifacts in the Seg\&Proj baseline are due to errors in stereo matching. For more qualitative examples, please see the supplementary material.
 
In general, large and more frequent classes (\emph{dirt, dirt-trail, grass} and \emph{bush}) are easier to predict than small, less frequent objects (\emph{tree} and \emph{rock}). This is expected since small objects are harder to localize, especially when they are far. \emph{Canopy} is also harder to localize due to its overhanging nature and occlusion. In terms of elevation, \bevnet{}-TA does not show notable improvement in the ground elevation than \bevnet{}. We hypothesize that it is due to the recurrent layer not being able to track the vertical movement of the vehicle well. We leave it as future work to improve this aspect. 

\begin{figure}[t]
    \centering
    \includegraphics[width=1\columnwidth]{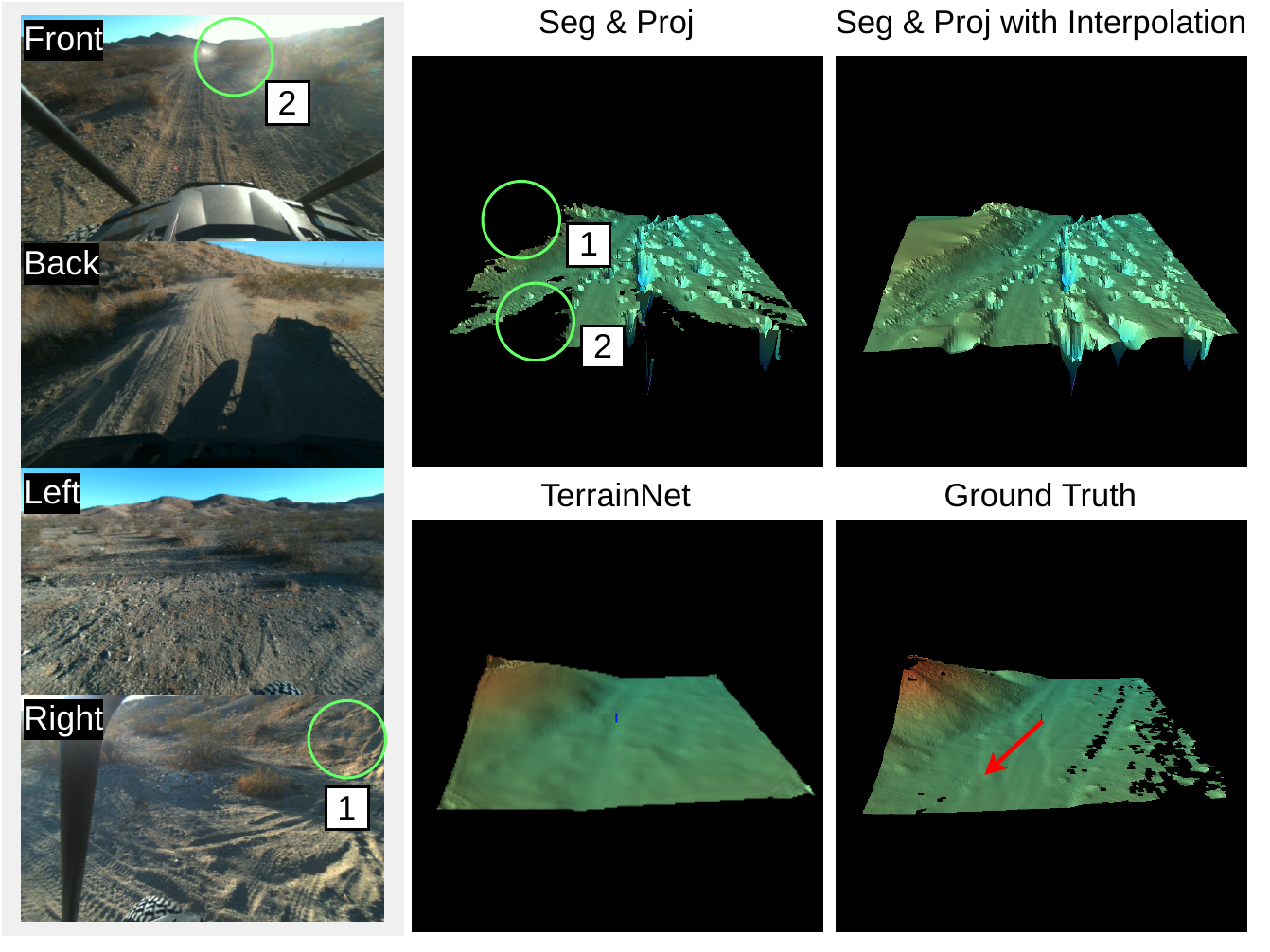}
    \caption{Comparing \bevnet{} to Seg\&Proj with interpolation. \bevnet{} is able to predict the non-visible hill crest (area 1) and the occluded flat ground (area 2) behind dense vegetation, whereas interpolating aggregated stereo depths predicts a flat hill and a bumpy ground. TerrainNet is also robust to artefacts in stereo depths.}
    \label{fig:interpolation}
    \figurespace
\end{figure}

\textbf{Stereo depth provides significant gains.} Models with stereo depths as additional input outperform their RGB-only counterparts by a large margin. For off-road environments, objects are randomly scattered, and they can have similar appearances but different sizes (e.g., small vs. large bushes). Thus, it is hard to estimate the depth from color information. This suggests that for off-road terrain perception, it is preferred to equip the robot with stereo cameras instead of monocular cameras. The performance improvement is usually worth the extra cost.

\textbf{Forward vs. backward projection.} SimpleBEV performs on par with other methods on \emph{dirt}, \emph{dirt-trail}, and \emph{grass}. These semantic classes cover a large area of the ground, so accurate depth estimation is not essential. But for objects such as \emph{bush}, \emph{canopy}, \emph{tree}, and \emph{rock}, SimpleBEV usually underperforms. SimpleBEV also performs poorly on elevation estimation, likely due to the low $z$-resolution of the grid. Increasing the grid resolution would linearly increase computation and memory consumption, which is not always feasible.

\textbf{Learning-based inpainting is better than interpolation.} In Figure~\ref{fig:interpolation}, we compare the predicted ground elevation map from Seg\&Proj with Navier-Stokes interpolation~\cite{bertalmio2001navier} and TerrainNet-TA. Only temporally aggregating the stereo depths creates an incomplete elevation map due to occlusion. A non-learning-based interpolation cannot leverage the context to predict the occluded areas, such as the hill crest and the flat ground behind dense vegetation. In contrast, \bevnet{} learns shape priors to predict occluded areas more accurately. TerrainNet is also less prone to the artefacts in the raw stereo depths.

\subsubsection{Inference Speed}
Inference speed is vital for fast and safe off-road driving. In Table~\ref{tab:quantitative-pseudo}, \bevnet{} is $7\times$ faster than LSS. The significant speedup comes from the direct projection of point features, which avoids the cost of projecting the features to every location on a depth ray. Adding stereo depth to the input has a minor impact on the speed because it only introduces a few extra convolution layers. Similarly, the temporal aggregation module has a relatively low overhead.

\subsection{Ablation Study}

\subsubsection{Architecture}
To understand the importance of each component in \bevnet{}, we perform an ablation study by training a few alternative models with some of the components disabled. We summarize the results in Table~\ref{tab:ablation} and have the following observations:
\begin{enumerate}
    \item Without soft quantization (\emph{``No soft quantization''}), the semantic mIoU degrades by 2 points, indicating the differentiable projection is crucial for correcting the end-to-end projection error from image space to BEV space. It has a small impact on elevation accuracy likely due to the terrain being naturally smooth.
    %\item The ``\emph{No z encoding}'' variant removes the terrain embedding network $\text{MLP}(x)$ and concatenates $z$ with image features directly. The resulting features are weighted and projected to the map. This variant causes a 1.6 points drop in semantic mIoU, but has no impact on ground elevation. \nathan{Which network does it remove, MLP($z'$) or MLP(concat($f_\text{sem},f_\text{elev}$)) ? When I was reading that section, I was confused why MLP($z'$) was necessary.}
    \item The ``\emph{No z}'' variant removes the elevation feature completely and only projects the image features. This results in a large error in ground elevation.
    \item Removing the depth completion network and only using the raw stereo depth (\emph{``No depth completion''}) degrades both semantic and elevation accuracy by a large margin.
    \item We also trained a model using the ground-truth depth for the projection (\emph{``Full model + GT depth''}). It has a sizeable improvement both in semantic mIoU (2 points) and elevation error.
\end{enumerate}

The results show that every design decision made in \bevnet{} is crucial for its performance. It also indicates that improving depth prediction accuracy is an effective way to improve the terrain model.

\subsubsection{Pseudo-labeling}
\label{sec:exp-pseudolabel}
Since we train a pseudo-labeling model to generate the semantic BEV labels for the unlabeled training frames, we compare this scheme with two alternatives that do not leverage pseudo-labeling: 1) using only labeled data (about 4500 training frames), and 2) using the whole training set but not applying the semantic loss to unlabeled frames (note that elevation loss is applied to every frame). 

Table~\ref{tab:ablation-data} compares the pseudo-labeling scheme with the two alternatives on the validation set (we do not apply pseudo-labeling to the validation set). Pseudo-labeling provides the largest gain in semantic prediction (3.7 points) but at a small cost in elevation accuracy. Tuning the weight between the semantic and elevation loss can potentially mitigate this problem, which we leave as future work.

\subsubsection{Map size}
\label{sec:map-size}
We choose $50~\mathrm{m}\times 50~\mathrm{m}$ maps for our main comparison because the cameras on the vehicle are tilted downwards such that the image content beyond $25~\mathrm{m}$ gets heavily compressed and all models struggle at larger maps. Nevertheless, our model can scale to $100~\mathrm{m}\times 100~\mathrm{m}$ maps better than Lift-Splat-Shoot. In Table~\ref{tab:map-size} we compare Lift-Splat-Shoot, \bevnet{}, and \bevnet{}-TA on $100~\mathrm{m}\times 100~\mathrm{m}$ maps with RGB-D inputs.

We see a similar trend in semantic accuracy: \bevnet{}-TA performs the best in semantic prediction. In terms of scalability, \bevnet{} is now $10\times$ faster than LSS. \bevnet{} projects each pixel to one map location, so its projection procedure is independent of map size. %In contrast, LSS projects each pixel to every map cell along a depth ray, making the computational cost linear to the map size.

\subsection{Planning with \bevnet{}}
\label{sec:planning}

To show that the output of \bevnet{} can be effectively used by a planner for off-road navigation, we also ran experiments with a planner in the loop.

\textbf{Planner.}
We use MPPI~\citep{williams2017information}, a sampling-based model predictive control algorithm, as our planner as it is effective for high-speed off-road driving~\citep{williams2016aggressive,williams2017information}. At each time step, we perform 3000 rollouts with a kinematic bicycle model and evaluate the cost of each rollout using the terrain features along the trajectory, as described below. Then, we compute the optimized control trajectory as a weighted average of the rollouts, with the weights computed from the rollouts' aforementioned costs.

\begin{table}[t]
  \centering
  \begin{tabular}{lcc}
    \toprule
    & mIoU $\uparrow$ & Ground Elevation (MAE) $\downarrow$\\
    \hline
    % 90.72k
    No soft quantization & 0.449 & 0.254\\
    % 96.39k
    % No z encoding & 0.464 & 0.246 \\
    % 105.8k
    No z & 0.451 & 0.628 \\
    % 83.16k
    No depth completion & 0.432 & 0.314 \\
    \hline
    Full model & 0.469 & 0.244 \\
    % 51.03k
    Full model + GT depth & 0.496 & 0.232 \\ 
    \bottomrule
  \end{tabular}
  \caption{Ablation study with \bevnet{} and RGB-D inputs. }
  \label{tab:ablation}
  \figurespace
\end{table}

\begin{table}[t]
  \centering
  \begin{tabular}{lcc}
    \toprule
    Training data & mIoU $\uparrow$ & Ground Elevation (MAE) $\downarrow$\\
    \hline
    Only manually labeled & 0.432 & 0.281\\
    Manually labeled + All elev. & 0.427 & \textbf{0.241} \\
    Pseudo labeled + All elev. & \textbf{0.469} & 0.244 \\
    \bottomrule
  \end{tabular}
  \caption{Effects of training data on the terrain modeling accuracy.}
  \label{tab:ablation-data}
  \figurespace
\end{table}

\begin{table}[t]
  \centering
  \begin{tabular}{lccc}
    \toprule
    Method & mIoU $\uparrow$ & Ground Elevation (MAE) $\downarrow$ & Time (ms) $\downarrow$\\
    \hline
    LSS & 0.399 & 0.628 & 402\\
    \bevnet{} & 0.419 & \textbf{0.586} & \textbf{39}\\
    \bevnet{}-TA & \textbf{0.464} & 0.622 & 42\\
    \bottomrule
  \end{tabular}
  \caption{Results on $100~\mathrm{m}\times 100~\mathrm{m}$ maps with RGB-D inputs.}
  \label{tab:map-size}
  \figurespace
\end{table}

\textbf{Cost function.}
The terrain cost $C$ of a trajectory $\tau$ is a sum of costs evaluated at each state: $C(\tau) =\sum_{t=1}^T G(s_t) + M(s_t)$, where $T$ is the planning horizon, $s_t$ is the planar vehicle state (location on the map, velocity, and heading) at time step $t$, $G(s)$ is the distance from state $s$ to some desired goal state, and $M(s)$ is the terrain cost at $s$.

\begin{figure*}[t]
    \centering
    \includegraphics[width=1\textwidth]{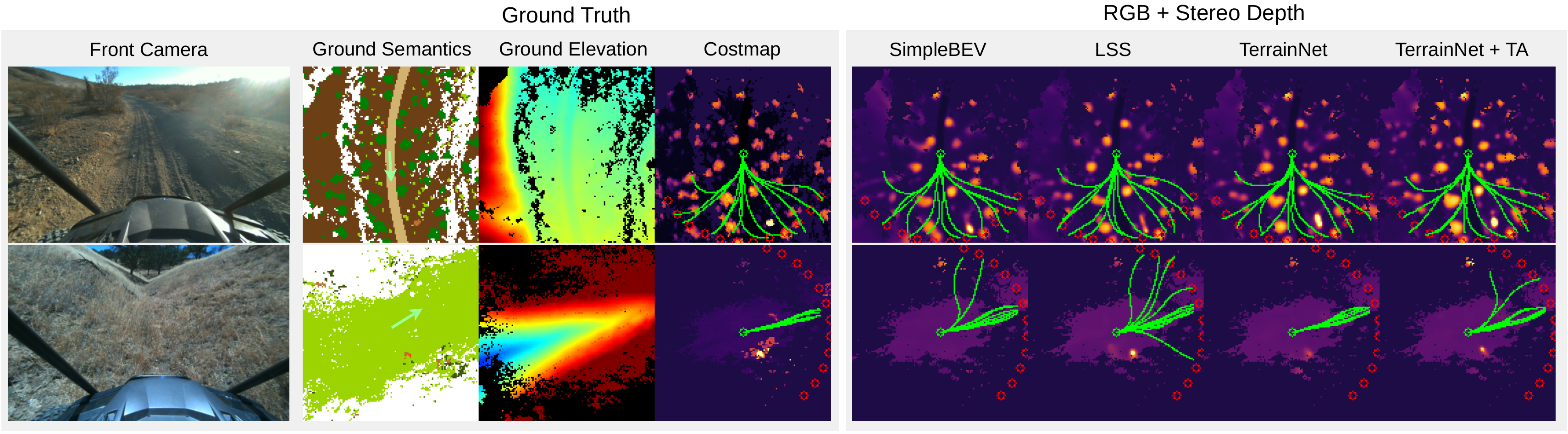}
    \caption{Visualization of MPPI control trajectories on two different terrains. Green arrow indicates vehicle's heading. Brighter areas correspond to higher costs. All models use RGB with stereo depth. \textbf{Top:} On-trail driving. The trajectories may go through non-lethal vegetation but do avoid the tree on the front left of the vehicle. \textbf{Bottom:} Driving in a steep valley. Due to the steep slopes, the trajectories should stay in the valley. We do not visualize the elevation cost here due to its dependence on vehicle heading.}
    \label{fig:compare-planning}
\end{figure*}

The terrain cost $M$ incorporates both semantic and geometric information based on the capabilities of the vehicle. There are several ways to do this. For the static comparative experiments in Figure~\ref{fig:compare-planning}, we compute it as follows:
\begin{align*}
M(s) &= c^\text{ground}_\text{semantic} + c^\text{ceiling}_\text{semantic} + c_\text{elevation} \\
c^\text{groud}_\text{semantic} &= (1+\alpha_1(h_\text{max}-h_\text{min})) \gamma_\text{ground}^\top \mathcal{C}_\text{ground}   \\
c^\text{ceiling}_\text{semantic} &= (1+\alpha_2(h_\text{min} + h_\text{clearance} - h_\text{ceiling})) \gamma_\text{ceiling}^\top \mathcal{C}_\text{ceiling} \\
c_\text{elevation} &= \beta_1\theta_\text{roll}^2 + \beta_2\theta_\text{pitch}^2,
\end{align*}
where $\alpha_1$, $\alpha_2$, $\beta_1$, and $\beta_2$ are scalar tuning parameters, $\gamma_\text{ground}$ and $\gamma_\text{ceiling}$ are tuning vectors of costs for each semantic class, and $\theta_\text{roll}$ and $\theta_\text{pitch}$ are the estimated roll and pitch angles of the vehicle based on the state $s$ and the elevation map from \bevnet{}. The height multiplier for semantic costs helps account for the fact that within any given semantic class, larger obstacles tend to be less traversable.
The elevation cost penalizes large pitch and roll angles. Large pitch angles may be too steep to ascend, whereas large roll angles may cause the vehicle to tip over. These angles depend on vehicle heading and can be computed using $h_\text{min}$ (or $h_\text{max}$) at the locations of the wheels.

\begin{figure*}
    \centering
    \includegraphics[width=1\textwidth]{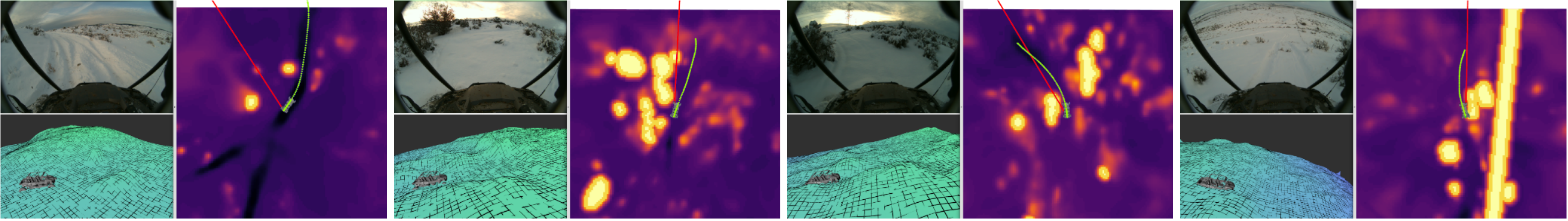}
    \caption{Qualitative results from a closed-loop real-world experiment. Each figure shows the front RGB camera (non-rectified), elevation map (showing $h_\text{max}$), and semantic costmap (showing $c_\text{semantic}$). Overlaid on the semantic costmap, the red line shows the direction of the goal point, and the green line shows the local plan. From the left: \textbf{(1) Driving on a trail with a hill to the left side.} Note that the trail is lower cost, so the local planner moves faster (as shown by the long green line). \textbf{(2) Driving off trail up a hill.} Note that the tall bushes on the left are very high cost. \textbf{(3) Driving uphill with many wheel slips on the snow.} The costmap and elevation map are ``smeared'' along the direction of travel, which highlights a limitation of temporal aggregation when odometry is poor. \textbf{(4) Approaching a steep downhill slope.} Although the slope itself is not yet directly visible, \bevnet{} predicts a steep elevation map. The local planner slows down to reduce the risk of rollover. (The high-cost straight line through the costmap is a manually GPS-specified keep-out zone to mark the location of a nearby fence.)}
    \label{fig:realworld}
    \figurespace
\end{figure*}

 For the real-world experiments~(Figure \ref{fig:realworld}), the terrain cost incorporates more physics:
 \begin{align}
 M(s) &= v^2 c^\text{ground}_\text{semantic} + C_\text{lethal}\mathbf{1}[c_\text{rollover} > \delta] \\
 c_\text{rollover} &= |\kappa v^2 + g \sin(\theta_\text{roll})| / \cos(\theta_\text{roll}),
 \end{align}
where $C_\text{lethal}$ is the lethal cost, $\mathbf{1}[\cdot]$ is the indicator function, $v$ is vehicle velocity, $\kappa$ is trajectory curvature, $g$ is the acceleration due to gravity, and $\delta$ is a tuning parameter depending on vehicle geometry.
 Multiplying the semantic cost $c^\text{ground}_\text{semantic}$ by $v^2$ reflects the fact that collisions are more dangerous at higher speeds. $c_\text{rollover}$ is the risk of rolling the vehicle based on speed, ground slope, and how sharply the vehicle is turning.

\begin{table}[t]
  \centering
  \begin{tabular}{lccc}
    \toprule
    Method & Avg. HD $\downarrow$ & Avg. CD $\downarrow$ & Time (ms) $\downarrow$\\
    % \hline
    % \textbf{RGB}&\\
    % SimpleBEV & 4.24\\
    % LSS & \textbf{3.97}\\
    % \bevnet{} & 4.14 \\
    \hline
    %\textbf{RGB + Stereo Depth}&\\
    SimpleBEV & 2.837 & 0.404 & 113\\
    LSS & 2.760 & 0.387 & 207\\
    \bevnet{} & \textbf{2.658} & \textbf{0.371} & \textbf{28}\\
    \bevnet{}-TA & 2.758 & 0.384 & 29\\ 
    \bottomrule
  \end{tabular}
  \caption{Planning evaluation with open-loop MPPI. Inputs are RGB-D. We compute the Average symmetric Hausdorff Distance and the Average Cost Difference between the optimized control trajectories on the predicted and the ground-truth costmaps.}
  \label{tab:planning}
  \figurespace
\end{table}


% \begin{table}[t]
%   \centering
%   \begin{tabular}{lccc}
%     \toprule
%     Method & Avg. Hausdorff Distance (AHD) $\downarrow$ & Cost Diff. $\downarrow$\\
%     \hline
%     % \textbf{RGB}&&\\
%     % SimpleBEV & 3.500 & 0.447\\
%     % LSS & \textbf{3.287} & \textbf{0.419}\\
%     % \bevnet{} & 3.460 & 0.440 \\
%     % \hline
%     \textbf{RGB + Stereo Depth}&&\\
%     SimpleBEV & 3.401 & 0.440\\
%     LSS & 3.158 & 0.383\\
%     \bevnet{} & \textbf{2.968} & \textbf{0.371}\\
%     \bevnet{}-TA & 3.118 & 0.376\\ 
%     \bottomrule
%   \end{tabular}
%   \caption{Planning evaluation with open-loop MPPI. We use the symmetric Hausdorff Distance as the metric. 
%   %\sasha{Is this Avg. HD? Caveat with HD: valid/low-cost trajectories may deviate largely from GT trajectories ex. going left instead of right around an obstacle. Another useful metric would be to take average difference in trajectory cost with the ground-truth.} \xiangyun{This would require hacking the MPPI code so that I can evaluate the trajectory cost from a baseline on the ground-truth costmap}
%   }
%   \label{tab:planning}
% \end{table}

\textbf{Static planning comparison.}
%\anqi{It is a little hard to see the difference between planning with \bevnet{} and other baselines, in terms of which performs better and which performs worse. I don't know how much value this comparison adds to the paper. Potentially we can just show planning results with \bevnet{} and \bevnet{}-TA only, showing that it can work with an actual planner. And also that the plan prodcued by \bevnet{} is closed to the one produced by ground truth maps.}
% 
First, we set up an experiment on the validation dataset by running the MPPI planner on the costmap computed from the output of \bevnet{} and other baselines (all with stereo inputs). The navigation task is as follows: we choose 12 waypoints in front of the vehicle at a distance of $25~\mathrm{m}$, spanning from $-60^\circ$ to $60^\circ$. Then, we run MPPI to plan a trajectory to each waypoint. %sasha{May be good to run multiple trajectories for each waypoint as well, to account for random sampling} \xiangyun{The current MPPI is too slow to run multiple trajectories for each waypoint...}. We compare the trajectories computed from the ground-truth and the predicted terrain maps. \sasha{Mention that, although approximate vehicle dynamics are used for simulation, this same model is used by MPPI to effectively predict vehicle behavior on the real system. It also serves as a scalable way to provide demonstrations for evaluation.}
Table~\ref{tab:planning} summarizes the results, and Figure~\ref{fig:compare-planning} visualizes example costmaps and trajectories. We observe the same trend where \bevnet{} with RGB-D inputs performs the best. One interesting observation is that \bevnet{}-TA performs worse than \bevnet{}. This is likely caused by the planner being more sensitive to certain terrain features than others, and that the vehicle mostly driving forward (making history less useful). Currently, for generality, we do not discriminate between different terrain classes, but it would be beneficial to weigh their influence on planning during training. We leave this as future work.

% \begin{enumerate}
%     \item With RGB input, LSS achieves lower AHD than \bevnet{}, even though LSS is worse on the semantic and elevation metrics. This is likely caused by the planner being more sensitive to certain terrain features than others (e.g., lethal objects affect planning more than non-lethal objects). Currently, for generality, we do not discriminate between different terrain classes, but it would be beneficial to weigh their influence on planning during training.
%     \item  Besides the reason mentioned above, we think another cause is that the vehicle mostly drives forward. Temporal aggregation is useful for retaining history, but it does not help much with predicting what is in front of the vehicle. Nevertheless, TA is still important when the vehicle gets stuck and needs to back up. 
%     %\xiangyun{Maybe I should evaluate backward planning as well.}
% \end{enumerate}

\textbf{Real-world navigation.}
\bevnet{} has been integrated with an off-road autonomous driving stack and tested in real-world environments. Limited by the season, we tested \bevnet{} in an off-road environment covered with deep snow and steep slopes. Since our dataset does not contain snow, we annotated 15 LiDAR scans with snow and finetuned our model with an updated ontology (including \emph{snow} and \emph{snow trail}). Figure~\ref{fig:realworld} (best viewed zoomed in) shows sample costmaps, elevation maps, and planner output from an autonomous run. The system completed a $1.1~\mathrm{km}$ run with a maximum speed of $7~\mathrm{m/s}$ (average of $3.2~\mathrm{m/s}$) with two human interventions. The interventions were mainly due to errors in odometry from wheel slips. \bevnet{} ran at $20~\mathrm{Hz}$ onboard the vehicle. Due to additional overhead from the system, \bevnet{} provided map updates at $10~\mathrm{Hz}$. See the website for more details. %We do notice that the map size could have been increased to provide a larger area. At speeds of 10 m/s, a $50\times 50$ map provides only 2.5 seconds' worth of semantic and geometric information in front of the vehicle.

\section{Discussion}
\textbf{LiDAR point density.} One question a reader may ask is whether increasing the number of LiDARs is sufficient for building a dense map at high speed. While more LiDARs increase the point density, it also increases the cost of the system. Compared to typical deployment scenarios where robots may only have one LiDAR, cameras provide much higher pixel density at a lower cost. Cameras also provide additional advantages such as higher reliability, stealth, and being less affected by small particles such as dust. Nonetheless, it would be interesting to compare LiDAR-based mapping and camera-based mapping in challenging environments.

\textbf{Limitations.} While \bevnet{} predicts high-fidelity terrain semantics and elevations, it has several limitations to be addressed:
\begin{itemize}
\item It may not generalize well to an environment that is substantially different from the training environments. This is common in learning-based systems. To improve the generalization, we can increase the diversity of the datasets and apply domain adaptation techniques.
\item It may miss fine-grained features such as small, lethal rocks. Due to hardware limitations, the cameras are not precisely synchronized with the LiDARs (note that LiDARs are used to generate training data). Hence, there exists a small but noisy misalignment between the projected camera features and the ground-truth maps. This makes it harder for the model to learn fine-grained details of the terrain. To address this, we can improve the sensor alignment and make the model less sensitive to the localization error of small objects.
\item \bevnet{} does not predict uncertainty. This can lead to dangerous behaviors, such as the model predicting the other side of a cliff as a ramp. We can incorporate uncertainty estimation or risk assessment as done by \citet{cai2022risk} and \citet{fan2021learning} to make the model risk-aware.
\end{itemize} 
\section{Conclusion}

In this paper, we designed and implemented \bevnet{}, the first off-road, camera-only perception system for joint BEV semantic and geometric terrain mapping. We demonstrated \bevnet{}'s accuracy and efficiency in terrain perception and successfully deployed the model as part of a navigation system for a variety of challenging off-road environments. In future work, we aim to include uncertainty estimates in the output, learn costmaps directly from expert demonstrations to remove manually engineered cost parameters, and more effectively leverage unlabeled data for learning.

\section*{Acknowledgments}
This research was developed with funding from the Defense Advanced Research Projects Agency (DARPA), ARL SARA CRA W911NF-20-2-0095, and ARL SARA CRA W911NF-21-2-0190.

\bibliographystyle{plainnat}
\bibliography{references}

\appendix

\subsection{Computing Terrain Representation}
This section provides more details about how we build the terrain representation.

\textbf{Point cloud aggregation.} Given the raw LiDAR packets and the vehicle's ego motion, we convert the LiDAR packets into LiDAR points with the rolling shutter effects corrected. Then, we run Google Cartographer \cite{cartographer} on the LiDAR points (with IMU information) to compute the gravity-aligned pose for each LiDAR scan. We aggregate the LiDAR scans using their poses to get an aggregated point cloud.

\textbf{Outlier removal.} LiDAR may produce false returns due to the presence of dust, snowflakes, and water. These false returns manifest as noise in the aggregated point cloud. We identify these outliers in two ways:
\begin{enumerate}
    \item Voxel filter. We voxelize the point cloud with a voxel size of 0.3m. We count the number of points in each voxel. If the number of points is smaller than a threshold (set to 5), we remove all the points in this voxel.
    \item Semantic filter. When we label the LiDAR points, we ask the labelers to label the outliers as an additional \emph{outlier} class. Our LiDAR segmentation network is trained with the additional outlier class. After we predict the pseudo labels for each point, we remove points classified as outliers.
\end{enumerate}

\subsection{Network Architecture}
\begin{figure}[t]
    \centering
    \includegraphics[width=1\columnwidth]{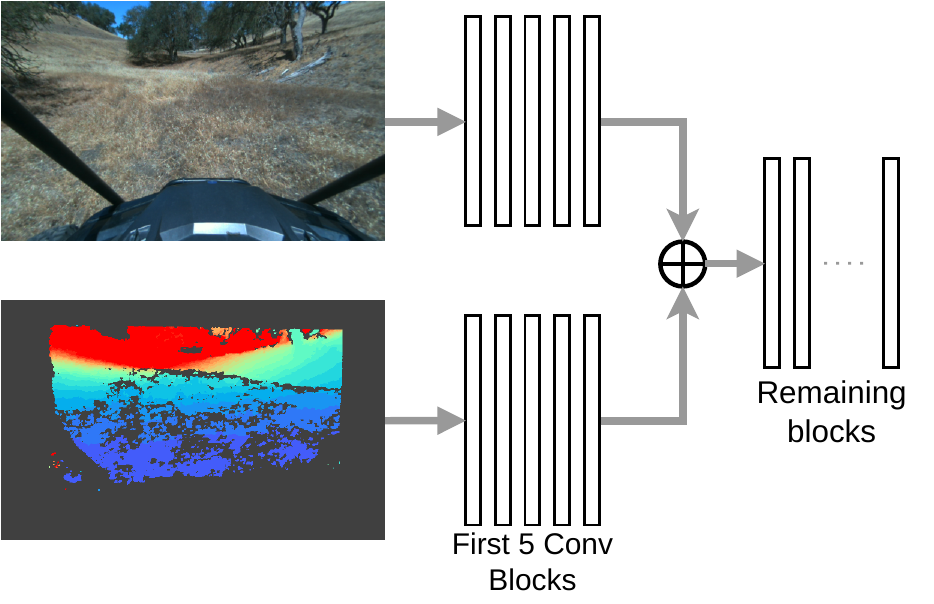}
    \caption{The image backbone uses separate branches to process the RGB and depth inputs, respectively. Then, the outputs of the two branches are summed and passed to the remaining blocks in the backbone.}
    \label{fig:depth_completion}
\end{figure}

\textbf{Backbone.} Our image backbone follows that of Lift-Splat-Shoot\cite{philion2020lift}. There are two main differences:
\begin{enumerate}
    \item For RGB-D inputs, we duplicate the first five convolutional blocks to process the RGB and depth images separately (Figure~\ref{fig:depth_completion}). This has a noticeable improvement over passing a 4-channel RGB-D image directly to the backbone.
    \item We do $8\times$ downsampling of the feature map and depth map (the original paper performs $16\times$ downsampling). This provides a good trade-off between speed and accuracy. Note that all the baselines are trained with the same image backbone with the same $8\times$ downsampling ratio.
\end{enumerate}
Our input image size is $512\times 320$. Hence, the image backbone produces a 472-channel feature map with a spatial dimension of $64\times 40$.

\textbf{Depth completion network.} The output of the image backbone is passed to a single convolution layer with 472 input channels and $D$ output channels with kernel size 3. $D$ is the number of desirable depth bins. For $50~\mathrm{m}\times 50~\mathrm{m}$ maps, $D=128$. For $100~\mathrm{m}\times 100~\mathrm{m}$ maps, $D=256$.

\textbf{Terrain embedding network.} For each back-projected point $(x, y, z)$, we first apply an MLP with layer sizes $(1, 64, 32)$ to $z$ to get the 32-dimensional elevation embedding $f_\text{elev}$. Then, we concatenate $f_\text{elev}$ with the 472-channel image feature $f_\text{sem}$ and apply another MLP with layer sizes $(504, 96)$ to get the 96-dimensional terrain embedding $f$. ReLU is applied after each MLP layer.

\textbf{Temporal aggregation layer.} The temporal aggregation layer is a single ConvGRU layer with 96 input channels, 96 output channels, and a kernel size of 1. To train TerrainNet-TA, we first train the non-recurrent version, and then insert the recurrent layer. We freeze the weights of the model except the recurrent layer.

\textbf{Inpainting network.} The inpainting net follows the architecture of Lift-Splat-Shoot's BEV encoder network. The main difference is that we create a separate decoding head consisting of an upsampling layer and two convolution layers to predict a specific terrain layer.

\subsection{Data Annotation}
\begin{figure*}[t]
    \centering
    \includegraphics[width=0.8\textwidth]{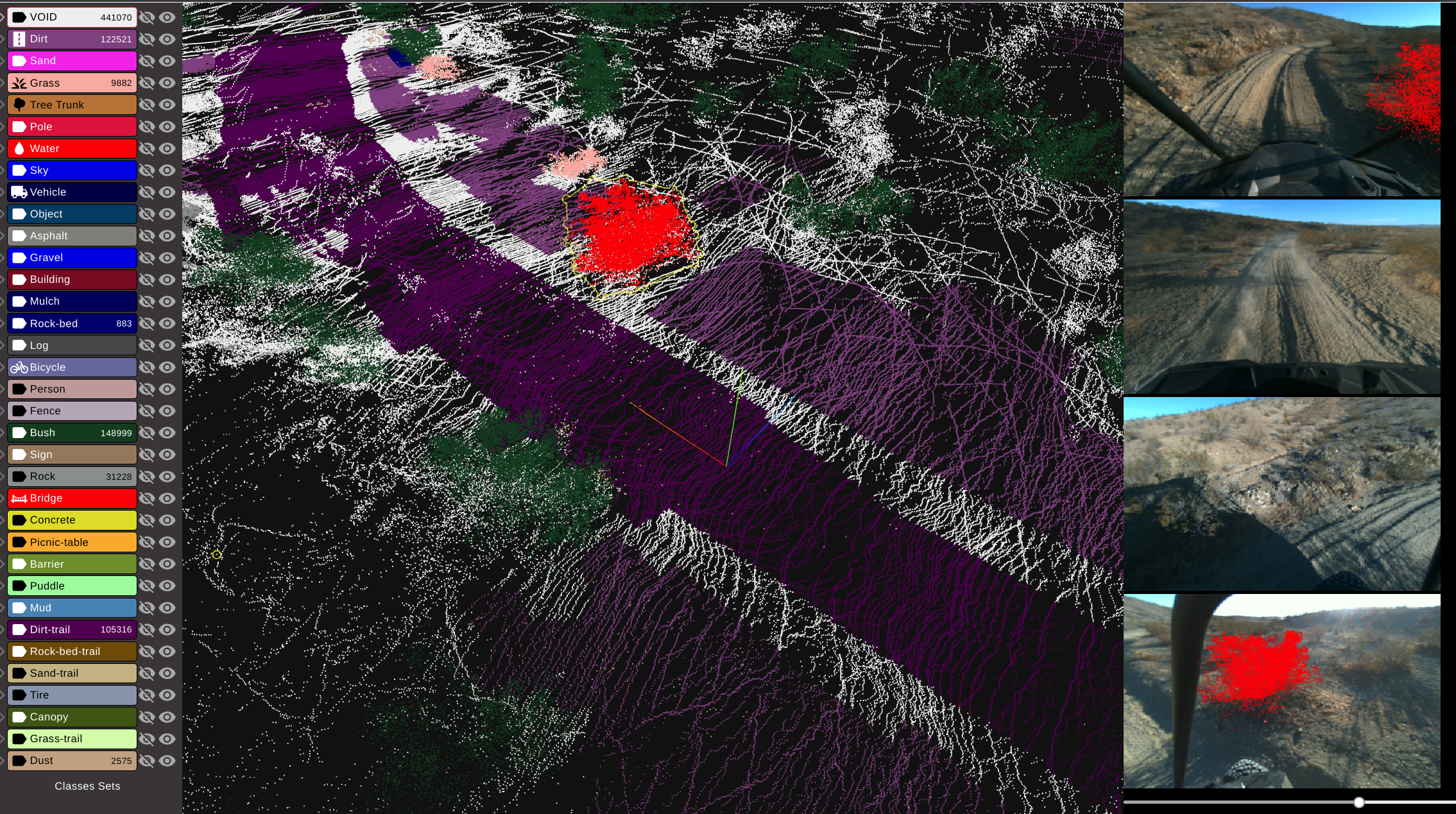}
    \caption{\textbf{Annotation tool.} Labelers can select points in the point cloud and see their corresponding image projections. Here we show a bush is highlighted both in the point cloud and in the bottom image. This helps the labelers to verify the semantic class of those points. Moreover, labelers only annotate points that they are confident about. We leave points that are difficult to identify unlabeled.}
    \label{fig:annotation-tool}
\end{figure*}

We create our data annotation tool (Figure~\ref{fig:annotation-tool}) to obtain ground truth semantically labeled point clouds. We aggregate 30 to 50 LiDAR scans for each selected frame to get a dense point cloud for labeling. We provide camera images for every LiDAR scan for reference so that a labeler can cross-reference to verify that the labels are correct.

\balance

\subsection{Additional dataset examples}
\begin{figure*}[t]
    \centering
    \includegraphics[width=1\textwidth]{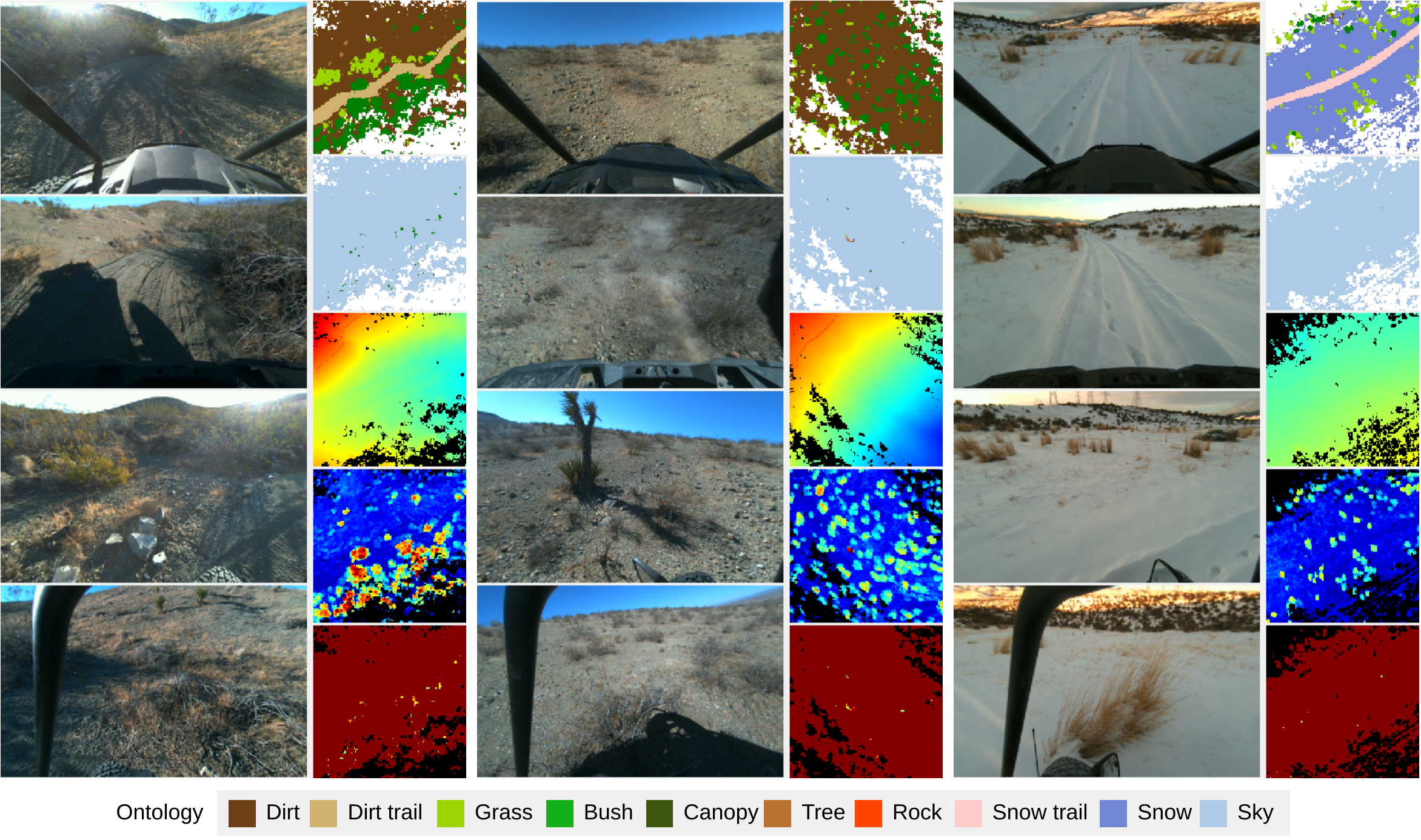}
    \caption{Additional dataset examples. These examples were collected at different locations and in different seasons. For the real-world experiments, we finetuned the model with annotated snow data (rightmost column).}
    \label{fig:more-examples}
\end{figure*}
Figure~\ref{fig:more-examples} shows additional dataset examples to highlight the diversity of the datasets. We collected our datasets from 3 distinctive geographic locations across the year.

\subsection{More qualitative results}
\begin{figure*}[t]
    \centering
    \includegraphics[width=1\textwidth]{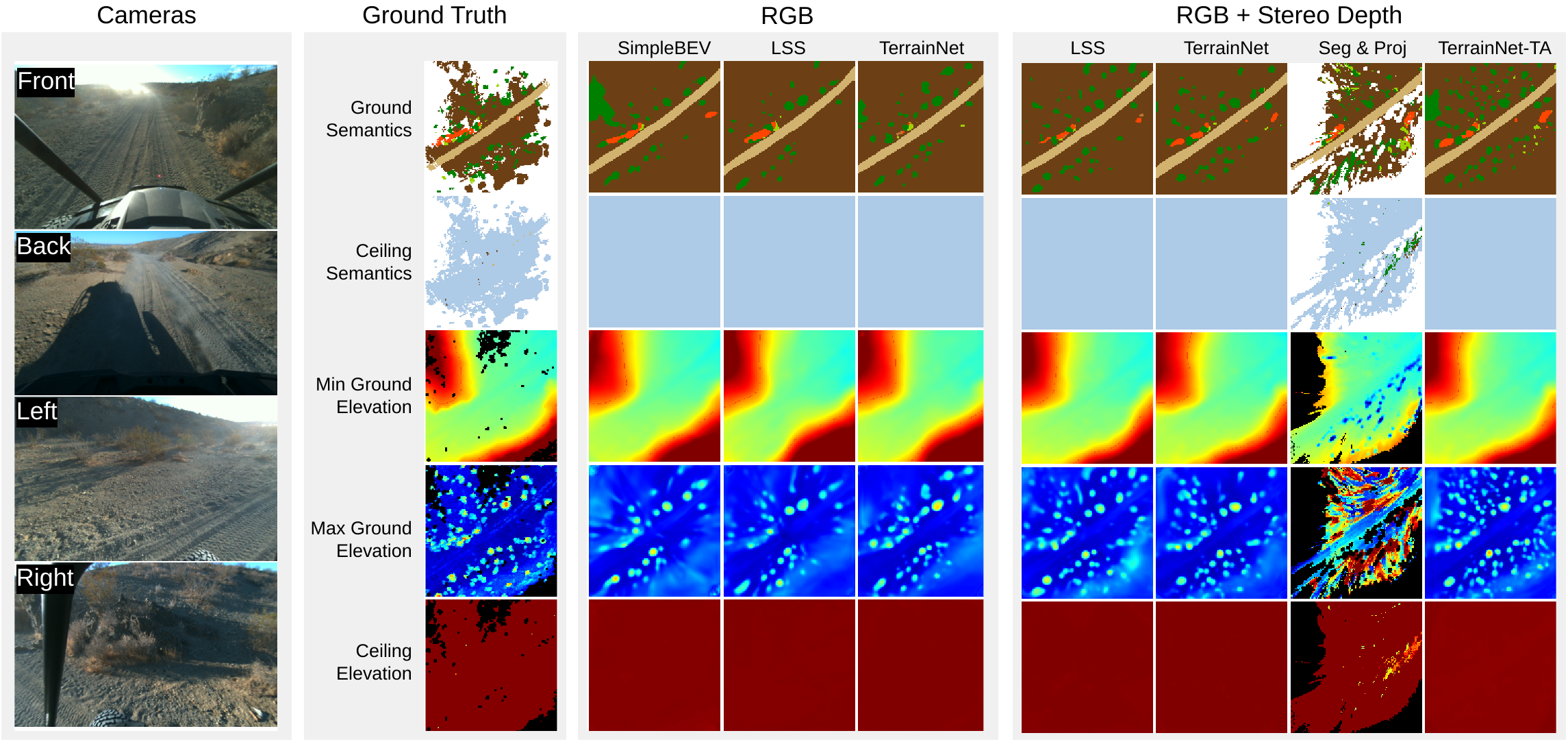}
    \includegraphics[width=1\textwidth]{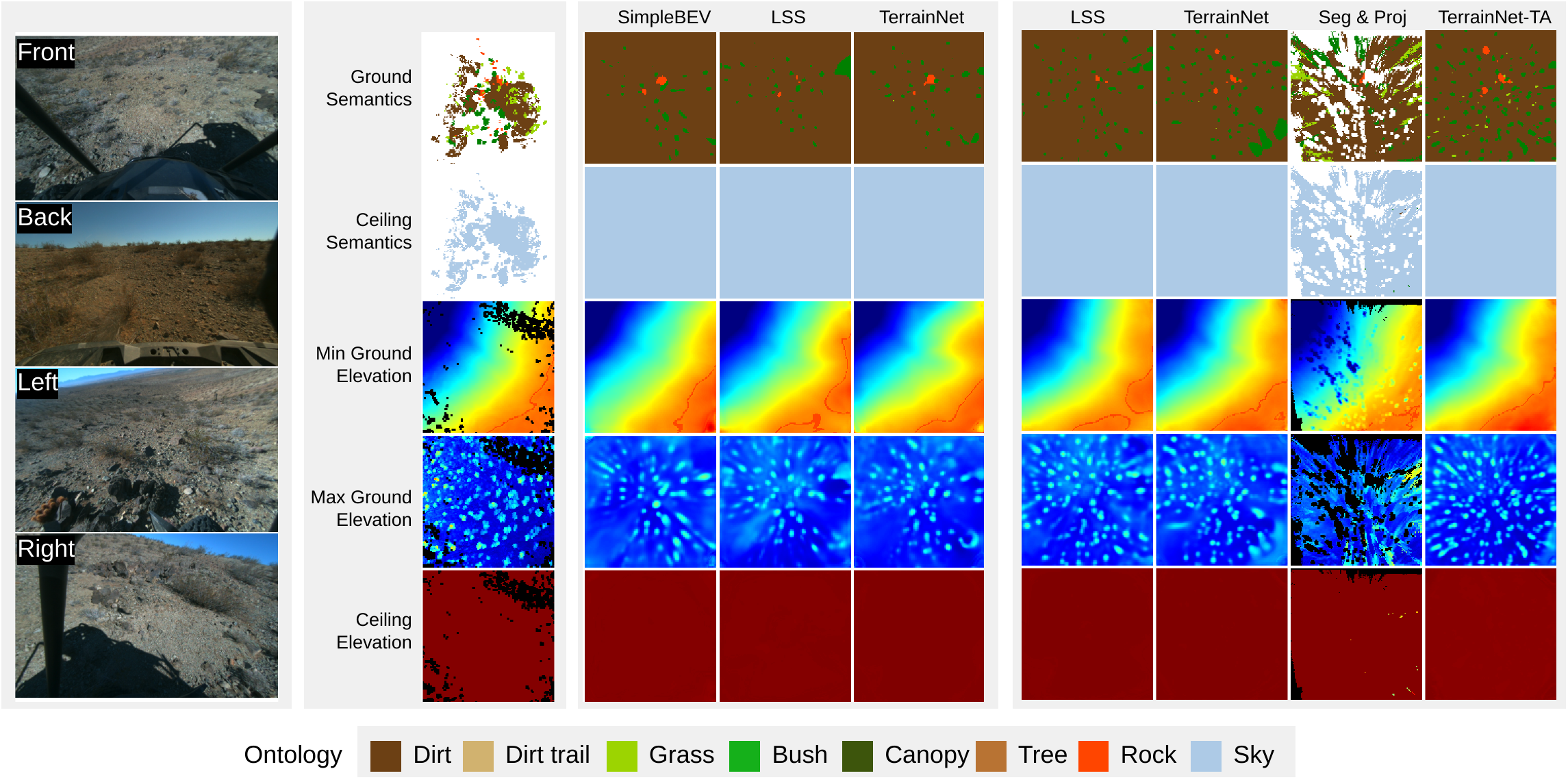}
    \caption{Comparing the predictions of different models. \textbf{Top:} on-trail run with rocks and bushes on the two sides of the trail. \textbf{Bottom:} off-trail run with scattered rocks and bushes. These are open terrains, so the ceiling semantics consists almost entirely of \emph{sky}.}
    \label{fig:more-pred-examples}
\end{figure*}
\begin{figure*}[t]
    \centering
    \includegraphics[width=0.7\textwidth]{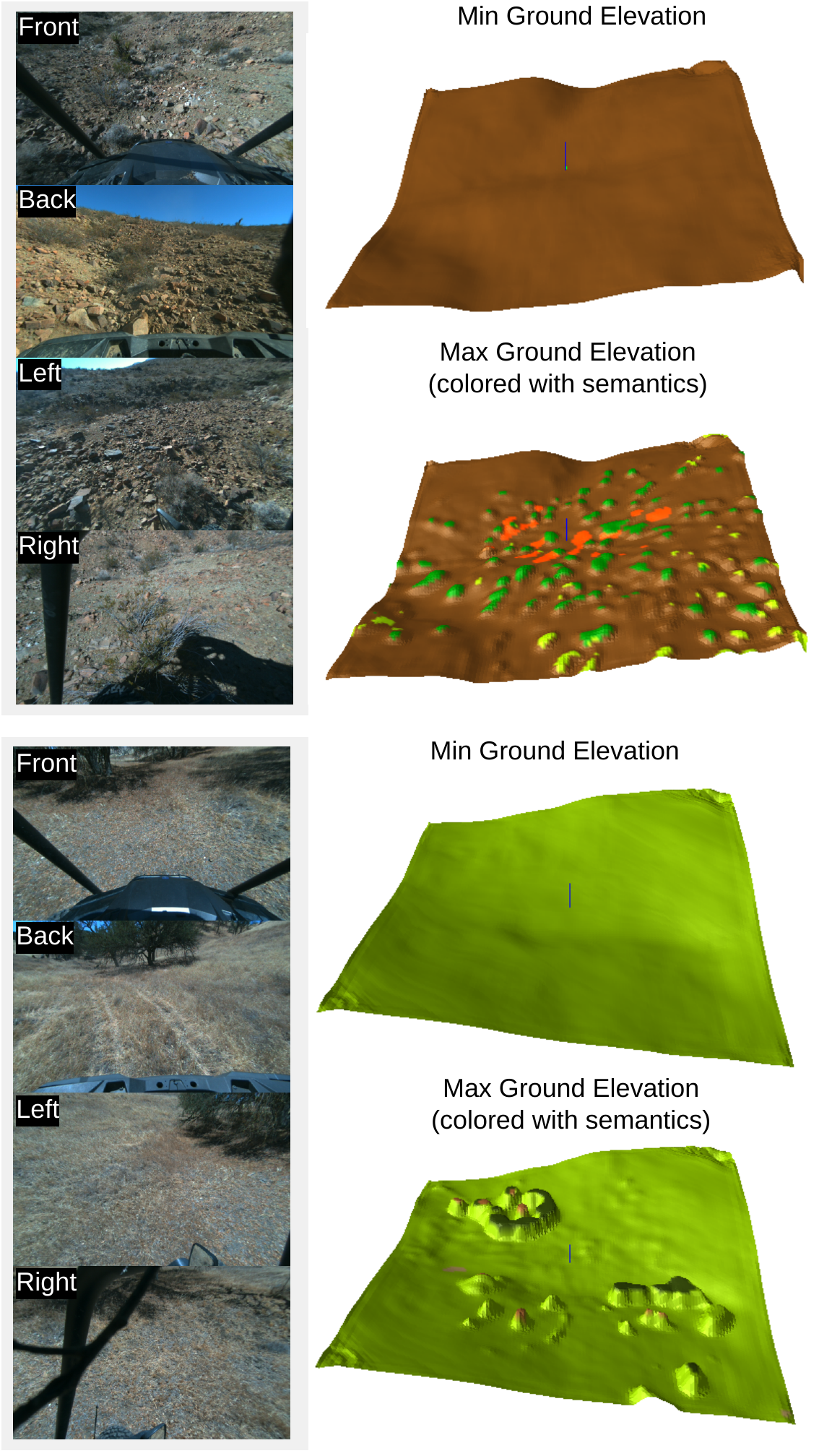}
    \caption{\textbf{Visualizations of the predicted 3D terrain model from \bevnet{}-TA}. We only show the min and max ground elevation here for clarity. Left: uneven terrain with scattered grass, bushes, and rocks. Right: a grass-covered valley with tall trees and low-hanging canopies.}
    \label{fig:more-3d-examples}
\end{figure*}
Figure~\ref{fig:more-pred-examples} shows other example predictions of \bevnet{} and other baselines on the validation sets. Figure~\ref{fig:more-3d-examples} visualizes different predicted 3D terrains of \bevnet{}-TA on the validation sets. For more details, including videos, please visit the website.

\end{document}